\documentclass[10pt,journal,compsoc]{IEEEtran}

\usepackage{booktabs} % For formal tables
\usepackage{microtype}
\usepackage[latin1]{inputenc}
\usepackage[OT2,T1]{fontenc}

\usepackage{amssymb}
\usepackage[linesnumbered,ruled,vlined]{algorithm2e}
\usepackage{algpseudocode}
\usepackage{bm}
\usepackage{amsmath}
\usepackage{amsthm}
\usepackage{graphicx}
\usepackage{url}
\usepackage{psfrag}
\usepackage{balance}
\usepackage{float}
\usepackage{multirow}
\usepackage{array}
\usepackage{enumitem}
\usepackage{hyperref}
\usepackage{pifont}
\usepackage{bbm}
\usepackage{caption}
\usepackage{subcaption}

\SetCommentSty{mycommfont}
\usepackage{diagbox}
\usepackage{tikz}
\usetikzlibrary{shapes.geometric,shapes.misc,positioning,backgrounds,fit}
\usepackage{scalerel}

\usepackage{pgfplots}
\usepackage{graphicx}

\newtheorem{thm}{Theorem}
\DeclareMathOperator*{\argmax}{argmax}
\DeclareMathOperator*{\argmin}{argmin}

\ifCLASSINFOpdf

\else

\fi

\begin{document}
	
\title{Fast Gumbel-Max Sketch and its Applications}

\author{Yuanming~Zhang, Pinghui~Wang, Yiyan~Qi, Kuankuan~Cheng, Junzhou~Zhao,\\  
	Guangjian~Tian and~Xiaohong~Guan
\IEEEcompsocitemizethanks{
\IEEEcompsocthanksitem 
An earlier conference version of this paper appeared at the Proceedings of The Web Conference 2020
~\cite{QiWZZTG20}. In this extended version, we extend the Gumbel-Max sketch's definition, which can be applied to more applications. Compared with the conference version, we also propose a more efficient method FastGM to compute the Gumbel-Max sketch and include new experiments for the application of weighted cardinality estimation.

\IEEEcompsocthanksitem 
Y. Zhang, P. Wang, Y. Qi, K. Cheng, and J. Zhao are with the MOE Key Laboratory for Intelligent
Networks and Network Security, Xi'an Jiaotong University, P.O. Box 1088,
No. 28, Xianning West Road, Xi'an, Shaanxi 710049, China.
E-mail: \{zhangyuanming, kuankuan.cheng\}@stu.xjtu.edu.cn, phwang@mail.xjtu.edu.cn, qiyiyan@idea.edu.cn, junzhou.zhao@xjtu.edu.cn.
\IEEEcompsocthanksitem X. Guan is with the MOE Key Laboratory for Intelligent Networks and
Network Security, Xi'an Jiaotong University, P.O. Box 1088, No. 28,
Xianning West Road, Xi'an, Shaanxi 710049, China 
and also
with the Center for Intelligent and Networked Systems, Tsinghua National
Lab for Information Science and Technology, Tsinghua University, Beijing
100084, China. E-mail: xhguan@mail.xjtu.edu.cn.
\IEEEcompsocthanksitem G. Tian is with Huawei Noah's Ark Lab, Hong Kong.
E-mail: Tian.Guangjian@huawei.com.
}% <-this % stops an unwanted space
\thanks{Corresponding author: Pinghui Wang}
\thanks{Manuscript received April 19, 2005; revised August 26, 2015.}}
\markboth{Journal of \LaTeX\ Class Files,~Vol.~14, No.~8, August~2015}%
{Shell \MakeLowercase{\textit{et al.}}: Bare Demo of IEEEtran.cls for Computer Society Journals}

\IEEEtitleabstractindextext{%
\begin{abstract}
The well-known Gumbel-Max Trick for sampling elements from a categorical distribution (or more generally a non-negative vector) and its variants have
been widely used in areas such as machine learning and information retrieval.
To sample a random element $i$ in proportion to its positive weight $v_i$,
the Gumbel-Max Trick first computes a Gumbel random variable $g_i$ for each positive weight element $i$, and then samples the element $i$ with the largest value of $g_i+\ln v_i$.
Recently,
applications including similarity estimation and weighted cardinality estimation require
to generate $k$ independent Gumbel-Max variables from high dimensional vectors.
However, it is computationally expensive for a large $k$ (e.g., hundreds or even thousands) when using the traditional Gumbel-Max Trick.
To solve this problem, we propose a novel algorithm, \emph{FastGM}, which reduces
the time complexity from $O(kn^+)$ to $O(k \ln k + n^+)$,
where $n^+$ is the number of positive elements in the vector of interest.
FastGM stops the procedure of Gumbel random variables computing for many elements, especially for those with small weights.
We perform experiments on a variety of real-world datasets
and the experimental results demonstrate that FastGM is orders of magnitude faster than state-of-the-art methods without sacrificing accuracy or incurring additional expenses.
\end{abstract}

% Note that keywords are not normally used for peerreview papers.
}

% make the title area
\maketitle

% To allow for easy dual compilation without having to reenter the
% abstract/keywords data, the \IEEEtitleabstractindextext text will
% not be used in maketitle, but will appear (i.e., to be "transported")
% here as \IEEEdisplaynontitleabstractindextext when the compsoc 
% or transmag modes are not selected <OR> if conference mode is selected 
% - because all conference papers position the abstract like regular
% papers do.
\IEEEdisplaynontitleabstractindextext
% \IEEEdisplaynontitleabstractindextext has no effect when using
% compsoc or transmag under a non-conference mode.

% For peer review papers, you can put extra information on the cover
% page as needed:
% \ifCLASSOPTIONpeerreview
% \begin{center} \bfseries EDICS Category: 3-BBND \end{center}
% \fi
%
% For peerreview papers, this IEEEtran command inserts a page break and
% creates the second title. It will be ignored for other modes.
\IEEEpeerreviewmaketitle

\section{Introduction} \label{sec:introduction}

The Gumbel-Max Trick~\cite{luce2012individual} is a popular technique for sampling elements
from a categorical distribution (or more generally a non-negative vector), which has been widely used in many areas.
Given a non-negative vector $\vec{v}=(v_1,\ldots,v_n)$, let
$N^+_{\vec{v}}\triangleq\{i\colon v_i > 0, i=1,\ldots,n\}$ be the set of
indices of positive elements in $\vec{v}$.
Then, the Gumbel-Max Trick computes a random variable
$s(\vec{v})$ as:
\begin{equation*}
  s(\vec{v}) \triangleq \argmax_{i\in N^+_{\vec{v}}} -\ln(-\ln a_i) + \ln v_i,
\end{equation*}
where
$a_i$ is a random variable drawn from the uniform distribution $\text{UNI}(0,1)$ independently and so $g_i=-\ln(-\ln a_i)$ is a Gumbel random variable.
Note that different vectors $\vec{v}$ should use the same set of variables $a_1, \ldots, a_n$ to guarantee consistency.
The distribution of random variable $s(\vec{v})$ is $P(s(\vec{v}) = i)=\frac{v_i}{\sum_{j=1}^n v_j}$.
Therefore, the Gumbel-Max Trick is popularly applied to sample an element from a high-dimensional non-negative vector $\vec{v}$ with probability proportional to the element's weight.

We call $s(\vec{v})$
and
$
  x(\vec{v}) = \max_{i\in N^+_{\vec{v}}} -\ln(-\ln a_i) + \ln v_i
$ 
as
Gumbel-ArgMax and Gumbel-Max variables of vector $\vec{v}$, respectively. 
In this paper, we define a Gumbel-Max  sketch of vector $\vec{v}$ as a vector of $k$ Gumbel-Max variables generated independently, i.e., $\vec{x}(\vec{v}) = (x_1(\vec{v}), \ldots, x_k(\vec{v}))$,
where 
$
  x_j(\vec{v}) = \max_{i\in N^+_{\vec{v}}} -\ln(-\ln a_{i,j}) + \ln v_i$, $j=1, \ldots, k$ and $a_{i,j}\sim \text{UNI}(0,1)$.
Similarly, we define  a Gumbel-ArgMax  sketch of vector $\vec{v}$ as a vector of $k$ Gumbel-ArgMax variables generated independently, i.e., $\vec{s}(\vec{v}) = (s_1(\vec{v}), \ldots, s_k(\vec{v}))$.
For simplicity, we also name the Gumbel-ArgMax sketch as the Gumbel-Max sketch when no confusion arises.
We observe that the Gumbel-Max sketch has been actually exploited for applications including probability Jaccard similarity estimation~\cite{yang2016poisketch,yang2017histosketch,yang2018d2,moulton2018maximally,yang2019nodesketch} and weighted cardinality estimation~\cite{lemieszAlgebraDataSketches2021}, while the authors of these works might be unconscious of this.

\textbf{Probability Jaccard Similarity Estimation.}
Similarity estimation lies at the core of many data mining and machine learning
applications, such as web duplicate detection
\cite{henzinger2006finding,manku2007detecting}, collaborate filtering
\cite{bachrach2009sketching} and association rule learning
\cite{MitzenmacherWWW14}.
To efficiently estimate the similarity between two vectors, several
algorithms~\cite{yang2016poisketch,yang2017histosketch,yang2018d2,moulton2018maximally}
compute $k$ random variables $-\frac{\ln a_{i,1}}{v_i}, \ldots, -\frac{\ln
	a_{i,k}}{v_i}$ for each positive element $v_i$ in $\vec{v}$, where $a_{i,1},
\ldots, a_{i,k}$ are independent random variables drawn from the uniform
distribution $\text{UNI}(0,1)$.
Then, these algorithms build a sketch of vector $\vec{v}$ consisting of $k$ registers, and each register records
$s_j(\vec{v})$ where
\begin{equation}\label{eq:sjv}
s_j(\vec{v}) = \argmin_{i\in N^+_{\vec{v}}} - \frac{\ln a_{i,j}}{v_i},
\quad 1 \le j \le k.
\end{equation}
We find that $s_j(\vec{v})$ is exactly a Gumbel-ArgMax variable of vector $\vec{v}$
as $\argmin_{i\in N^+_{\vec{v}}} - \frac{\ln a_{i,j}}{v_i} = \argmax_{i\in
	N^+_{\vec{v}}} \ln v_i - \ln (- \ln a_{i,j})$.
Let $\mathbbm{1}(x)$ be an indicator function.
Yang et al.~\cite{yang2016poisketch,yang2017histosketch,yang2018d2} use
$\frac{1}{k}\sum_{j=1}^k \mathbbm{1}(s_j(\vec{u}) = s_j(\vec{v}))$ to estimate the
\emph{weighted Jaccard similarity} of two non-negative vectors $\vec{u}$ and
$\vec{v}$ which is defined by
\[
\mathcal{J_W}(\vec{u},\vec{v})\triangleq
\frac{\sum_{i=1}^n \min\{u_i, v_i\}}{\sum_{i=1}^n \max\{u_i, v_i\}}.
\]
Recently, Moulton et al.~\cite{moulton2018maximally} prove that the expectation of
estimate $\frac{1}{k}\sum_1^k \mathbbm{1}(s_j(\vec{u}) = s_j(\vec{v}))$ actually
equals the \emph{probability Jaccard similarity}, which is defined by
\[
\mathcal{J_P}(\vec{u}, \vec{v})\triangleq
\sum_{i\in N^+_{\vec{v},\vec{u}}}
\frac{1}{\sum_{l=1}^n \max\left(\frac{u_l}{u_i}, \frac{v_l}{v_i}\right)}.
\]

Here, $N^+_{\vec{v},\vec{u}}\triangleq\{i\colon v_i > 0\wedge u_i > 0, i=1,
\ldots, n\}$ is the set of indices of positive elements in both $\vec{v}$ and
$\vec{u}$.
Compared with the weighted Jaccard similarity $\mathcal{J_W}$, Moulton et al. demonstrate that the probability Jaccard similarity $\mathcal{J_P}$ is
scale-invariant and more sensitive to changes in vectors.
Moreover, each function $s_j(\vec{v})$ maps similar vectors to the same value
with a high probability.
Therefore, similar to regular locality-sensitive hashing (LSH)
schemes~\cite{GionisPVLDB1999,Broder2000,Charikar2002similarity}, one can use these Gumbel-Max sketches to build an LSH index for fast similarity search in a large
dataset, which is capable to search similar vectors for any query vector in
sub-linear time.

\textbf{Weighted Cardinality  Estimation.} 
Given a sequence  $\Pi = o_1 o_2\cdots$, where $o_j\in \{1, \ldots, n\}$ represents an object (e.g., a string) and each object $i\in\{1, \ldots, n\}$ may appear more than once.
Each object $i$ has a positive weight $v_i$.
Let $N_\Pi$ be the set of objects that occurred in $\Pi$.
Then, the weighted cardinality of $\Pi$ is defined as $c_\Pi=\sum_{i\in N_{\Pi}} v_i$.
Take a SQL query "SELECT DISTINCT CompanyNames FROM Orders" as an instance.
The size (in bytes) of the query result is a sum weighted by string length over the "CompanyNames".
Besides the cardinality of a single sequence, sometimes, the data of interest consists of multiple sequences distributed over different locations and the target is to estimate the sum of all \emph{\textbf{unique}} occurred objects' weights using as few resources (including memory space, computation time, and communication cost) as possible.
The state-of-the-art method of weighted cardinality estimation is Lemiesz's sketch~\cite{lemieszAlgebraDataSketches2021}.
Let $\vec{v}^{\Pi}=(v_1^{\Pi}, \ldots, v_n^{\Pi})$ be the underlying vector of sequence $\Pi$. That is, each element $v_i^{\Pi}$, $i=1,\ldots,n$ equals $v_i$ (i.e., the weight of object $i$) when object $i$ occurs in sequence $\Pi$ (i.e., $i\in N_{\Pi}$) and 0 otherwise.
Lemiesz computes a sketch $\vec{y}(\vec{v}^{\Pi}) = (y_1(\vec{v}^{\Pi}), \ldots, y_k(\vec{v}^{\Pi}))$, and $y_j(\vec{v}^{\Pi})$ is defined as:
\begin{equation}\label{eq:yjpi}
y_j(\vec{v}^{\Pi}) = \min_{i\in N_{\Pi}} - \frac{\ln a_{i,j}}{v_i},
\quad 1 \le j \le k,
\end{equation}
where all variables $a_{i,j}\sim \text{UNI}(0,1)$ are independent with each other. 
Then, we easily find that the Gumbel-Max variable $x_j(\vec{v}^{\Pi}) = \max_{i\in N_{\Pi}} -\ln(-\ln a_{i,j}) + \ln v_i = -\ln y_j(\vec{v}^{\Pi})$.
Therefore, Lemiesz's sketch is a variant of the Gumbel-Max sketch.
It is easy to find that each $y_j(\vec{v}^{\Pi})$ follows the exponential distribution $\text{EXP}(c_\Pi)$ because 
$
P(y_j(\vec{v}^{\Pi}) \geq t)=\prod_{i\in  N_{\Pi}}P\left(-\frac{\ln a_{i,j}}{v_i}\geq t\right)
=e^{-c_\Pi t}.
$
Therefore, the sum $\sum^k_{j=1}y_j(\vec{v}^{\Pi})$ follows the gamma distribution $\Gamma(k, c_\Pi)$.
Based on the above observation, Lemiesz's algorithm estimates the weighted cardinality $c_\Pi$ as $\frac{k-1}{\sum^k_{j=1}y_j(\vec{v}^{\Pi})}$.
The proposed sketch is mergeable, 
which facilitates efficiently estimating the weighted cardinality of a sequence represented as a joint of different sequences $\Pi_1, \ldots, \Pi_d$.
Specifically, given the sketches of all $\Pi_1, \ldots, \Pi_d$, the sketch of the joint sequence $\Pi$ is computed as:
\[
y_j(\vec{v}^{\Pi}) = \min_{l=1, \ldots, d} y_j(\vec{v}^{\Pi_l}).
\]
Therefore, we only need to compute and gather the sketches of all sequences $\Pi_1, \ldots, \Pi_d$ together, which significantly reduces the memory usage and communication cost.

To compute the Gumbel-Max sketches of a large collection of vectors
(e.g., bag-of-words representations of documents),
the straightforward method first instantiates variables $a_{i,1}, \ldots, a_{i,k}$ from $\text{UNI}(0,1)$ for each index $i=1, \ldots, n$.
Then, for each non-negative vector $\vec{v}$, 
it enumerates each $i\in N^+_{\vec{v}}$ and computes $-\frac{\ln a_{i,1}}{v_i}, \ldots, -\frac{\ln a_{i,k}}{v_i}$.
The above method requires memory space $O(nk)$ to store all $\left[a_{i,j}\right]_{1\le i\le n, 1\le j\le k}$, 
and time complexity $O(k n^+_{\vec{v}})$ to obtain the Gumbel-Max sketch of each vector $\vec{v}$,
where $n^+_{\vec{v}}=
|N^+_{\vec{v}}|$ is the cardinality of set $N^+_{\vec{v}}$.
We note that $k$ is usually set to be hundreds or even thousands~\cite{moulton2018maximally,yang2019nodesketch,buchnik2019self}. Therefore, the straightforward method costs a huge amount of memory space and time when the vector of interest has a large dimension, e.g., $n=10^9$.
To reduce the memory cost,
one can easily use hash techniques or random number generators with specific seeds (e.g., consistent random number generation methods in~\cite{ShrivastavaUAI2014,ShrivastavaICML2014,ShrivastavaICML2017}) to generate each of $a_{i,1}, \ldots, a_{i,k}$ on the fly,
which does not require to calculate and store variables $\left[a_{i,j}\right]_{1\le i\le n, 1\le j\le k}$ in memory.

\begin{figure}[t]
	\centering
    \includegraphics{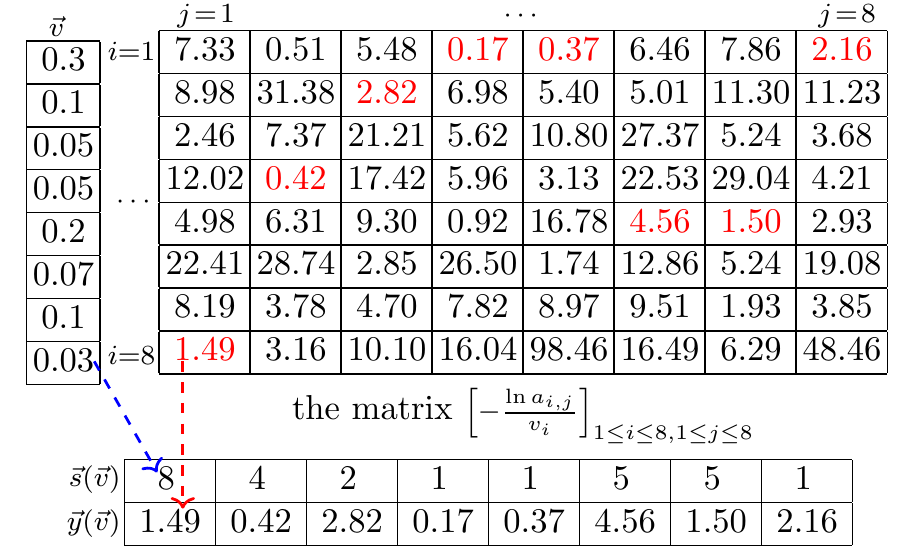}
	\caption{
	An example of computing the Gumbel-Max sketch $\vec{y}(\vec{v})$ and $\vec{s}(\vec{v})$ of length $k=8$ for a vector 
	$\vec{v}$,
        of which 
		elements  
		$y_j(\vec{v})$ and $s_j(\vec{v})$  respectively record the smallest element value (i.e., the red one) and its index in the $j$-th column of the matrix.
	}		
	\label{fig:exmple}
\end{figure}

\begin{figure}[t]
	\centering
     \includegraphics{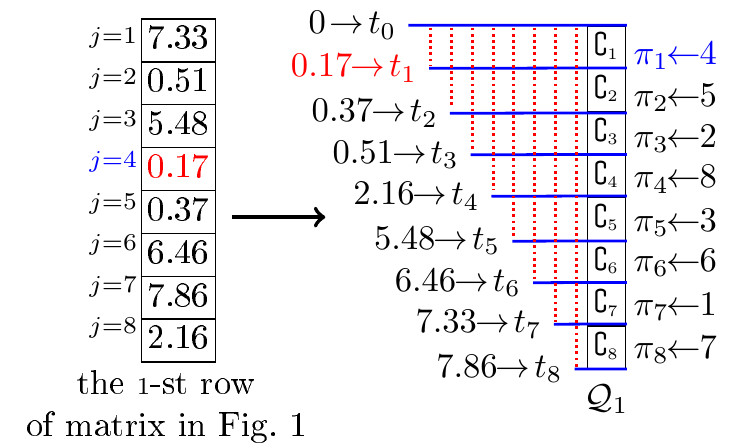}
	\caption{
	An example of building a queue $\mathcal{Q}_1$ from $8$ random variables in the $1$-st row of matrix 
	in Fig.~\ref{fig:exmple}. We use $\complement$ to represent a customer in the queue.
    }
	\label{fig:queue_model}
\end{figure}

To address the computational challenge, in this paper, we propose a novel method \emph{FastGM} to fast compute a
Gumbel-Max sketch,
which reduces the time complexity of computing the sketch from $O(k  n^+_{\vec{v}})$ to $O(k \ln k + n^+_{\vec{v}})$.
From the example in Fig.~\ref{fig:exmple},
we find two interesting observations in computing the Gumbel-Max sketch in a straightforward way:
1) 
The $k$ variables $-\frac{\ln a_{i,1}}{v_i}, \ldots, - \frac{\ln a_{i,k}}{v_i}$ of a relatively larger element $v_i$ in the $\vec{v}$ are more likely to be 
the Gumbel-Max variables,
such as the $v_1=0.3$ and $v_5=0.2$; 
2) Each Gumbel-Max variable occurs as one of an element $v_i$'s top minimal variables,
e.g., all three Gumbel-Max variables (red ones in the first row) appeared in $v_1$'s Top-4 minimal variables.
Therefore, we prioritize the generation order of all $kn$ variables and reduce the number of generated variables for computing the Gumbel-Max sketch.  
The basic idea behind our FastGM can be summarized as follows.
For each element $v_i>0$ in $\vec{v}$, we generate $k$ random variables
$-\frac{\ln a_{i,1}}{v_i}, \ldots, - \frac{\ln a_{i,k}}{v_i}$ in
ascending order.
As shown in Fig.~\ref{fig:queue_model},
we can generate a sequence of $k$ tuples $\left(t_{i,1}, \pi_{i,1}\right), \ldots, \left(t_{i,k}, \pi_{i,k}\right)$,
where $t_{i,j}=-\frac{\ln a_{i, \pi_{i,j}}}{v_i}$, $t_{i,1}< \cdots < t_{i,k}$ and $\pi_{i,j}=i_j$,
$(i_1, \ldots, i_k)$ is a random permutation of integers $1,\ldots, k$.
When we are able to compute the Gumbel-Max sketch of $\vec{v}$ by obtaining the $k$ random variables in ascending order,
it is easy to find once the current obtained $t_{i,j}$ in tuple $\left(t_{i,j}, \pi_{i,j}\right)$ is larger than all elements in the $\vec{y}(\vec{v})$, there is no need to obtain the following tuples $\left(t_{i,j+1}, \pi_{i,j+1}\right),\cdots, \left(t_{i,k}, \pi_{i,k}\right)$ because they have no chance to change the Gumbel-Max sketch of $\vec{v}$.
Based on this property, we model the procedure of computing the Gumbel-Max sketch as a queuing model with $k$-servers and $n$-queues of different arrival rates.
Specifically, each queue has $k$ customers and each customer randomly selects a server.
A server $j=1, \ldots, k$
just serves the first arrived customer and ignores the other arrived customers.
In addition, a server $j$ only records the arrival time and the queue number (i.e., from which queue the customer comes) of its first arrived customer as $y_j(\vec{v})$ and $s_j(\vec{v})$, $j=1, \ldots, k$ respectively,
which have the same probability distributions as the variables $y_j(\vec{v})$ and $s_j(\vec{v})$ defined in Eq.~(\ref{eq:yjpi}) and Eq.~(\ref{eq:sjv}).
When each of the servers has processed its first arrived customer,
we close all queues and obtain the Gumbel-Max sketch $y_j(\vec{v})$ and $s_j(\vec{v})$ of $\vec{v}$. 
Based on the above model, we propose \emph{FastGM} to fast compute the Gumbel-Max sketch.
We summarize our main contributions as:
\begin{itemize}[leftmargin=*]
\item We introduce a simple queuing model to interpret the procedure of
  computing the Gumbel-Max sketch of vector $\vec{v}$.
  Using this stochastic process model, we propose a novel algorithm, called
  FastGM, to reduce the time complexity of computing the Gumbel-Max sketch
  $(s_1(\vec{v}), \ldots, s_k(\vec{v}))$ and $(y_1(\vec{v}), \ldots, y_k(\vec{v}))$ from $O(n^+_{\vec{v}}k)$ to $O(k \ln k + n^+_{\vec{v}})$,
  which is achieved by avoiding calculating all $k$ variables $- \frac{\ln
    a_{i,1}}{v_i}, \ldots, - \frac{\ln a_{i,k}}{v_i}$ for each $i \in
  N^+_{\vec{v}}$.
\item We conduct experiments on a variety of real-world datasets for applications including probability Jaccard similarity estimation and weighted cardinality estimation.
  The experimental results demonstrate that our method FastGM is orders of
  magnitude faster than the state-of-the-art methods without incurring any
  additional cost.
\end{itemize}

The rest of this paper is organized as follows.
Section~\ref{sec:method} and Section~\ref{sec:extension} present our method FastGM and its extension Stream-FastGM for non-streaming and streaming settings respectively.
The performance evaluation and testing results are presented in
Section~\ref{sec:results}. Section~\ref{sec:related} summarizes related work.
Concluding remarks then follow.

\section{Our Method FastGM}\label{sec:method}

In this section, we first introduce the basic idea behind our method FastGM through a simple example.
Then, we elaborate on FastGM in detail and discuss its space and time complexities.

\subsection{Basic Idea}\label{sec:basicidea}
In Fig.~\ref{fig:exmple}, we provide an example of generating a Gumbel-Max sketch of a vector $\vec{v} =(0.3, 0.1, 0.05, 0.05, 0.2, 0.07, 0.1, 0.03)$ to illustrate our basic idea, where we have $n=8$ and $k=8$.
Note that we aim to fast compute each $y_j(\vec{v})$ and $s_j(\vec{v})$, where $y_j(\vec{v}) = \min_{1\le i\le 8} -\frac{\ln a_{i,j}}{v_i}$ and $s_j(\vec{v}) = \argmin_{1\le i\le 8} -\frac{\ln a_{i,j}}{v_i}$, $1\le j\le 8$,
i.e., in each column $j$ of matrix $\left[-\frac{\ln a_{i,j}}{v_i}\right]_{1\le i\le 8, 1\le j\le 8}$ $y_j(\vec{v})$ records the minimum element and $s_j(\vec{v})$ records the index of this element.
We generate matrix $\left[-\frac{\ln a_{i,j}}{v_i}\right]_{1\le i\le 8, 1\le j\le 8}$ based on the traditional Gumbel-Max Trick and mark the minimum element (i.e., the red one indicating the Gumbel-Max variable) in each column $j$.
We find that Gumbel-Max variables tend to equal index $i$ with large weight $v_i$.
For example, among the values of all Gumbel-Max variables $s_1(\vec{v}), \ldots, s_{8}(\vec{v})$, index $1$ with $v_1 = 0.3$ appears 3 times, while index $3$ with $v_3 = 0.05$ never occurs.
Based on the above observations, 
we prioritize the generation order of the $64$ variables $-\frac{\ln a_{i,j}}{v_i}, 1\le i\le 8, 1\le j\le 8$
according to their values.
We first respectively select $R_i$ smallest variables from each row $i$ to compute the Gumbel-Max sketch,
where $R_i$ is proportional to the weight $v_i$.
The total number $R=R_1+\ldots+R_n$ of variables from all rows is computed as $R=k\ln k$.
This is the expected number of generated variables
before each column $j$ has at least one variable. 
To some extent, it is similar to the \emph{Coupon collector's problem}~\cite{motwani19953}.
Specifically, in the example of Fig.~\ref{fig:exmple}, we have $R=17=\lceil8\times\ln8\rceil$,
and each $R_i$ is computed as 
$R_i = \lceil R v^*_i \rceil$, where $\vec{v}^*=(v_1^*, \ldots, v_8^*)$ is the normalized vector of $\vec{v}$.
We have $R_1=6$, $R_2 = 2$, $R_3 = 1$, $R_4 = 1$, $R_5 = 4$, $R_6 = 2$, $R_7 = 2$, and $R_8 = 1$.
Meanwhile, we find that each Gumbel-Max variable occurs as one of a row $i$'s Top-$R_i$ minimal elements.
For example, the two Gumbel-Max variables occurring in the 5-th row are all among the Top-$R_5$ (i.e., Top-$4$) minimal elements.
Moreover,  we easily observe that $k$ random variables  $-\frac{\ln a_{i, 1}} {v_i}, \ldots, -\frac{\ln a_{i, k}} {v_i}$ in each row indeed are $k$ independent random variables follow the exponential distribution $\text{EXP}(v_i)$. 
Therefore, we can generate these $k$ variables in ascending order 
by exploiting 
the distribution of the order statistics of exponential random variables.
Based on the above insights, we derive our method FastGM.
As the example in Fig.~\ref{fig:queue_model}, for each row, we construct such a queue $\mathcal{Q}_i$ with arrival rate $v_i$ for the $k$ variables drawn from the distribution $\text{EXP}(v_i)$ according to their values.
Then, we first compute the variables that are in the front of the queues or 
in the queues with large arrival rates $v_i$, 
because they are smaller ones among all $kn$ variables and are more likely to become the Gumbel-Max variables.
Moreover, we early stop a queue when its remaining variables have no 
chance to be the Gumbel-Max variables.
Also take Fig.~\ref{fig:exmple} as an example.
Compared with the straightforward method computing all $nk=64$ random variables,
we compute $s_1(\vec{v}), \ldots, s_k(\vec{v})$ by only obtaining Top-$R_i$ minimal elements of each row $i$,
which significantly reduces the computation cost to around $\sum_{i=1}^8 R_i = 19$.
In summary, our method FastGM efficiently computes the Gumbel-Max sketch $\vec{y}(\vec{v})$ and $\vec{s}(\vec{v})$ of vector $\vec{v}$ through managing the number and order of variables $-\frac{\ln a_{i,j}}{v_i}$ belonging to different elements $v_i$.
Specifically, we aim to \emph{fast search} and compute those variables that have a high probability to become the elements of the Gumbel-Max sketch, and \emph{fast prune} variables have no chance to be an element in the Gumbel-Max sketch. In the following, when no confusion arises, we simply write $s_j(\vec{v})$, $y_j(\vec{v})$ and $n^+_{\vec{v}}$ as $s_j$, $y_j$ and $n^+$ respectively.

\subsection{Fast Gumbel-Max Sketch Generation}\label{sec:fastgm}
Our FastGM first constructs a queue $\mathcal{Q}_i$ for variables in each row as shown in Fig.~\ref{fig:queue_model}.
Based on this, we propose two modules: 
\textbf{FastSearch} for efficiently searching small variables in each queue, 
and \textbf{FastPrune} for pruning overlarge queues that cannot contribute to the Gumbel-Max sketch.
Before introducing our FastGM in detail, we first illustrate how to build a queue $\mathcal{Q}_i$ and
model the procedure of computing the Gumbel-Max sketch from another perspective via a \emph{Queuing Model  with $k$-servers and $n$-queues.}

\begin{figure*}[t]
	\centering
	\includegraphics{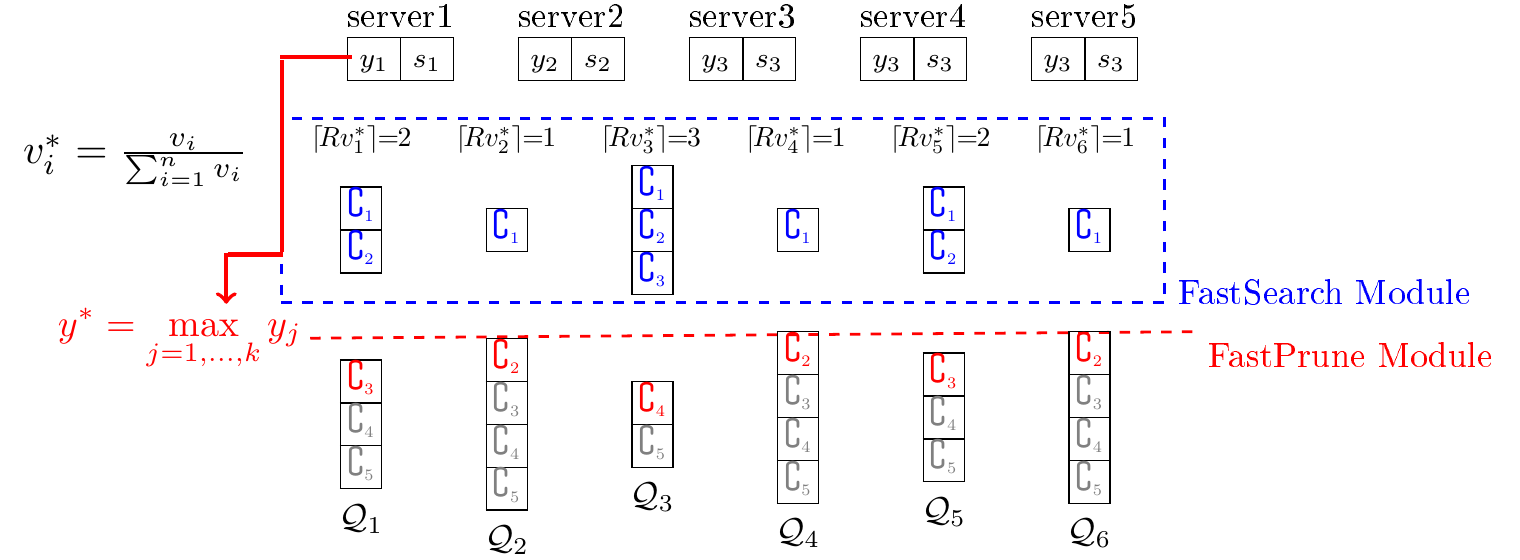}
	\caption{Illustration of our FastGM. The FastSearch module first selects a number of customers $\complement$ from each queue as candidates (i.e., the blue ones) and uses their arrival time to compute the $y^*=\max_{j=1,\ldots,k} y_j$. Then, the FastPrune module closes a queue according to the arrival time of the next customer (i.e., the red one) in each queue compared with the value of $y^*$, if $y^*$ is smaller, we close the queue for the rest of customers (i.e., the gray ones), which have no chance to be the first customers for any servers.}
	\label{fig:overview}
\end{figure*}

\textbf{Queuing Model with $k$-servers and $n$-queues.} 
In Fig.~\ref{fig:queue_model}, we show how to construct a queue $\mathcal{Q}_i$ where $k$ random variables $-\frac{\ln a_{i,1}}{v_i}, \cdots,-\frac{\ln a_{i,k}}{v_i}$ of $v_i$ are sorted in ascending order.
For simplicity, we define a variable $b_{i,j}$ as:
\begin{equation}\label{eq:gumbelvariable}
b_{i,j} = \frac{-\ln a_{i,j}}{v_i}, \qquad  i=1,\cdots,n \quad j=1,\cdots,k .
\end{equation}
We easily observe that $b_{i, 1}, \ldots, b_{i, k}$
are equivalent to $k$ independent random variables generated according to the exponential distribution $\text{EXP}(v_i)$.
Let $b_{i, (1)} < b_{i, (2)}< \ldots< b_{i, (k)}$ be the order statistics corresponding to variables $b_{i, 1}, \ldots, b_{i, k}$.
We construct each queue $\mathcal{Q}_i$ with  
$k$ customers $\complement_{i,j}$ whose arrival time $t_{i,j} \triangleq b_{i, (j)}, j=1,\ldots,k$ 
and each customer randomly selects a server $j$.
Specifically, we generate a sequence of $k$ tuples $\left(b_{i, (1)}, i_1\right), \ldots, \left(b_{i, (k)}, i_k\right)$, where 
$(i_1, \ldots, i_k)$ is a random permutation of integers $1,\ldots, k$ and denotes the server sequence
randomly selected by customers in 
a queue $\mathcal{Q}_i$.
It is easy to observe that values (resp. positions) of element $v_i$'s $k$ variables are the customers' arrival time (resp. selected servers) in the queue $\mathcal{Q}_i$.
Accordingly, we assign arrival time $t_{i,j}\triangleq b_{i, (j)}$ and selected server $\pi_{i,j}\triangleq i_j$ for each customer, i.e., $\left(t_{i,j}, \pi_{i,j}\right)\triangleq \left(b_{i, (j)}, i_j\right), j=1,\ldots,k$.
Note that, $k$ variables $b_{i, 1}, \ldots, b_{i, k}$ follow $\text{EXP}(v_i)$ with a rate parameter $v_i$. 
Therefore, customers in queue $\mathcal{Q}_i$ also arrive at this rate $v_i$. 
As shown in Fig.~\ref{fig:overview}, based on the built queues, the procedure of computing the Gumbel-Max sketch can be modeled as a \emph{Queuing Model with $k$-servers and $n$-queues}, where each server only 
serves the first arrived customer (i.e., records this customer's arrival time and the index of queue $i$, same as Eq.~(\ref{eq:yjpi}) and Eq.~(\ref{eq:sjv})).
Then, we naturally have the following two fundamental questions for the design of FastGM:

\noindent\textbf{Question 1.} How to fast search customers with the smallest arrival time to become candidates for the servers from these $N^+_{\vec{v}}$ queues?

\noindent\textbf{Question 2.} How to early stop a queue $\mathcal{Q}_i$, $i\in N^+_{\vec{v}}$?

We first discuss Question 1.
We note that customers of different queues $\mathcal{Q}_i$ arrive at different rates $v_i$.
Recall the example in Fig.~\ref{fig:exmple}, the basic idea behind the following technique is that queue $\mathcal{Q}_i$ with a high rate $v_i$ 
is more likely to produce customers with the smallest arrival time (i.e. Gumbel-Max variables).
Especially, when $z$ customers have arrived, let $t_{i,z}$ denote the arrival time of the $z$-th customer in queue $\mathcal{Q}_i$.
We find that $t_{i, z}$ can be represented as the sum of $z$ identically distributed exponential random variables with mean $\frac{1}{k v_i}$ (another perspective can be found in paper \cite{QiWZZTG20}).
Therefore, the expectation and variance of variable $t_{i, z}$ are computed as
\begin{equation}\label{eq:xizexpvar}
\mathbb{E}(t_{i, z}) = \frac{z}{k v_i}, \quad \text{Var}(t_{i, z}) = \frac{z}{k^2 v_i^2}.
\end{equation}
We easily find that $\mathbb{E}(t_{i, z})$ is $l$ times smaller than $\mathbb{E}(t_{j, z})$ when $v_i$ is $l$ times larger than $v_j$.

To obtain the first $R$ customers of the joint of all queues $\mathcal{Q}_i$, $i\in N^+_{\vec{v}}$, we let each queue $\mathcal{Q}_i$ release
$R_i = \lceil R v_i^* \rceil$ customers, where $\vec{v}^*$ is the normalized vector of $\vec{v}$.
Then, we have
$R \approx \sum_{i=1}^n R_i$.
For all $i\in N^+_{\vec{v}}$, their $t_{i, R_i}$ approximately have the same expectation.
\begin{equation}\label{eq:exiri}
\mathbb{E}(t_{i, R_i}\mid R) \approx \frac{R}{k \sum_{j=1}^n v_j}, \quad i\in N^+_{\vec{v}}.
\end{equation}
Therefore, the $R$ customers with the smallest arrival time are expected to be released.

Next, we discuss Question 2, which is inspired by the generation of ascending-order random variables.
For an element with index $j$ in the Gumbel-Max Sketch, 
we use two registers $y_j$ and $s_j$ to keep track of information on the customer with the smallest arrival time among all the released customers selected by server $j$,
where $y_j$ records the %variable's value 
customer's arrival time
and $s_j$ records the %variable's origin, 
index of the queue this customer comes from,
i.e., $\mathcal{Q}_{s_j}$.
When all servers $1,\ldots,k$ have been selected by at least one customer, 
we let $y^*$ keep track of the maximum value of $y_1, \ldots, y_k$, i.e.,
\[
y^* = \max_{j=1,\ldots,k} y_j.
\]
Then, we can stop queues $\mathcal{Q}_i$
when a customer coming from $\mathcal{Q}_i$ has an arrival time larger than $y^*$
because the arrival time of the subsequent customers from $\mathcal{Q}_i$ is also larger than $y^*$, which will not change any $y_1, \ldots, y_k$ and $s_1, \ldots, s_k$.

Based on the above two discussions, we develop our method FastGM to fast generate a $k$-length Gumbel-Max sketch with  $\vec{s}_{\vec{v}}=(s_1, \ldots, s_k)$ and $\vec{y}_{\vec{v}}=(y_1, \ldots, y_k)$ of any non-negative vector $\vec{v}$.
As shown in Fig.~\ref{fig:overview}, 
FastGM consists of two modules: FastSearch and FastPrune.
FastSearch is designed to quickly search 
customers with the smallest arrival time coming from all queues $\mathcal{Q}_1,\ldots,\mathcal{Q}_n$
and check whether all servers $1,\ldots,k$ have received at least one appointment from customers (i.e., selected by at least one customer).
When no servers are unreserved, 
we start the FastPrune module to close each queue
$\mathcal{Q}_i$, $i\in N^+_{\vec{v}}$.
We perform the procedure of FastPrune because
following customers coming from $\mathcal{Q}_i$
may also have an arrival time
smaller than $y^*$ and the 
customers may become the first arrived customers for some servers $j$ and change the values of $y_j$ and 
$s_j$ after the procedure of FastSearch.
Before we introduce these two modules in detail,
we first elaborate on the method of generating exponential random variables in ascending order, which is a building block for both modules.

\noindent\textbf{$\bullet$ Generating Ascending Exponential Random Variables}:
Next we detail how to sequentially generate $k$ random variables $b_{i,1}, \ldots, b_{i,k}$ in ascending order 
for each positive element $v_i$ of vector $\vec{v}$ (Lines 9-14 and Lines 24-29 in Algorithm~\ref{alg:FastGM-static}).
As we mentioned,
these $b_{i, 1}, \ldots, b_{i, k}$ 
are random variables according to the exponential distribution $\text{EXP}(v_i)$, i.e.,
\begin{equation}\label{eq:bij_exp}
	b_{i, j} \sim \text{EXP}(v_i), \quad j=1, \ldots, k.
\end{equation}
Let $b_{i, (1)} < b_{i, (2)}< \ldots< b_{i, (k)}$ be the order statistics corresponding to variables $b_{i, 1}, \ldots, b_{i, k}$.
Alfr\'{e} R\'{e}nyi~\cite{orderstatistic} 
observes that each $b_{i,(z)}$, $z=1,\ldots,k$ 
satisfies
\begin{equation}\label{eq:biz_dis}
{\displaystyle b_{i,(z)}\sim\frac{1}{v_i}\left(\sum _{n=1}^{z}{\frac {-\ln u_{i,n}}{k-n+1}}\right)},
\end{equation}
where all variables $u_{i,1}, \ldots, u_{i,z} \sim \text{UNI}(0,1)$ are independent random variables.
Note that $-\ln u_{i,n}$ is an $\text{EXP}(1)$ distributed random variable.
Therefore, one easily obtains the following equation:
\begin{equation}\label{eq:biz_biz-1}
	{b_{i,(z)}} - {b_{i,(z-1)}} \sim \frac{1}{v_i}\left({\frac {-\ln u_{i,z}}{k-z+1}}\right), \quad 2\leq z \leq k.
\end{equation}

Based on the above observation, 
we generate the order statistics $b_{i, (1)} ,\cdots, b_{i, (k)}$ for each element $v_i$ of vector $\vec{v}$ in an iterative way as:
\[
b_{i,(z)}\gets b_{i,(z-1)} + \frac{1}{v_i}\left(\frac{-\ln u_{i,z}}{k-z+1}\right), \quad 1\leq z \leq k,
\]
where $b_{i, (0)}=0$.
In addition, we use the Fisher-Yates shuffle~\cite{fisher1953statistical} (Lines 11-12 and lines 26-27 in Algorithm~\ref{alg:FastGM-static})
to iteratively produce a random permutation $i_1, \ldots, i_k$ for integers $1,\ldots, k$.
For an array $(\pi_{i,1}, \ldots, \pi_{i,k})$ with elements $(1, \ldots, k)$, 
in each step $z$, $1\leq z \leq k$, this method randomly selects a number $i_z$ from ${z, z+1,\ldots,k}$ and 
swaps the two elements in the array with indices $z$ and $i_z$.
To build the queue $\mathcal{Q}_i$, we assign $b_{i,(z)}$ and the element with index $z$ in the array to 
the arrival time and selected server of $z$-th customer, respectively, i.e.,
\[
t_{i,z} \gets b_{i,(z)}, \qquad i_z \gets \pi_{i,z}.
\]
We easily find that $k$ variables $b_{i, (1)} ,\cdots, b_{i, (k)}$ shuffled by the random permutation $i_1, \ldots, i_k$
have the same distribution as the variables $b_{i,1},\cdots, b_{i,k}$ generated in a direct manner.

\noindent\textbf{$\bullet$ FastSearch Module}: This module fast searches
customers with the smallest arrival time, and consists of the following steps:
\begin{itemize}[leftmargin=1.1cm]
	
\item[\textbf{Step 1:}] Iterate on each $i\in N^+_{\vec{v}}$ and repeat to generate $\lceil R v_i^*\rceil$ 
exponential variables (i.e., the arrival time of customers) in ascending order (Lines 9-14 in Algorithm~\ref{alg:FastGM-static}). 
Meanwhile, each server $j$ uses registers $y_j$ and $s_j$ to keep track of information of 
the first arrived customer,
where $y_j$ records the customer's arrival time and $s_j$ records the index of the queue
where the customer comes from (Lines~\ref{ln:syj_st}-\ref{ln:syj_ed} in
Algorithm~\ref{alg:FastGM-static});

\item[\textbf{Step 2:}] If there remain any unreserved servers, we increase $R$ by $\Delta$ and then repeat Step 1.
Otherwise, we stop the FastSearch procedure.
\end{itemize}
For simplicity, we set the parameter $\Delta = k$.
In our experiments, we find that the value of $\Delta$ has a small effect on the performance of FastGM.

\noindent\textbf{$\bullet$  FastPrune Module}:
When all servers $1, \ldots, k$ have been
selected by at least one customer among all the released customers.
We start the FastPrune module, which mainly consists of the following two steps:
\begin{itemize}[leftmargin=1.1cm]
\item[\textbf{Step 1.}] Compute $y^* = \max_{j=1,\ldots,k} y_j$.

\item[\textbf{Step 2.}] For each $\mathcal{Q}_i$, $i\in N^+_{\vec{v}}$,
we repeat to compute the next customer's arrival time (Lines 24-29 in Algorithm~\ref{alg:FastGM-static}).
Once a customer's arrival time is larger than $y^*$, we stop 
releasing customers from queue $\mathcal{Q}_i$(Lines 30-32 in Algorithm~\ref{alg:FastGM-static}).
As we mentioned, variables $y_j$ and $s_j$ keep track of information of 
the first arrived customer.
Therefore, $y_1, \ldots, y_k$ and $s_1, \ldots, s_k$ may also be updated by 
receiving new appointments from newly released customers with arrival times smaller than $y^*$
at this step (Lines 33-36 in Algorithm~\ref{alg:FastGM-static}).
Therefore, $y^*$ may also decrease with the number of released customers, 
which accelerates the termination of all queues $\mathcal{Q}_i$, $i\in N^+_{\vec{v}}$.
\end{itemize}

\subsection{Mergeability}
For some applications, 
the dataset of interest $\Pi$ is distributed over multiple sites.
Suppose that there are $r$ sites, each site $i=1, \ldots, r$ holds a sub-dataset $\Pi_i$.
Each site $i$ can compute the Gumbel-Max sketch $(\vec{s}^{(i)}, \vec{y}^{(i)})$ of its set $N^{(i)}$, which is the set of objects appearing in $\Pi_i$.
Here set $N^{(i)}$ can be easily represented as a weighted vector
following the weighted cardinality estimation discussed in Section~\ref{sec:introduction} and its Gumbel-Max sketch $(\vec{s}^{(i)}, \vec{y}^{(i)})$ can be computed based on our method FastGM.
A central site can collect all sites' sketches $(\vec{s}^{(1)}, \vec{y}^{(1)}), \ldots, (\vec{s}^{(r)}, \vec{y}^{(r)})$
and then use them to compute the Gumbel-Max sketch $(\vec{s}^{\cup}, \vec{y}^{\cup})$ of the union set $N^{(1)}\cup\cdots\cup N^{(r)}$.
For the sketch of union set, each element $y_i^{\cup}$, $j=1, \ldots, k$ of $\vec{y}^{\cup}$ is computed as
$y_j^{\cup} = \min_{i=1,\ldots,r}  y_j^{(i)}$,
where $y_j^{(i)}$ is the $j$-th element of vector $\vec{y}^{(i)}$.
Each element $s_i^{\cup}$ of $\vec{s}^{\cup}$ is computed as $s_j^{\cup} = s_j^{(i^*)}$, where $i^* = \argmin_{i=1,\ldots,r} y_j^{(i)}$ and  $s_j^{(i^*)}$ is the $j$-th element of vector $\vec{s}^{(i^*)}$.
At last, the weighted cardinality of dataset $\Pi$ can be estimated from the above Gumbel-Max sketch $(\vec{s}, \vec{y})$.

\subsection{Error Analysis}
As aforementioned, the parts $\vec{s}(\vec{v})$ and $\vec{y}(\vec{v})$ of Gumbel-Max sketches produced by FastGM are equivalent to the sketches proposed in \cite{moulton2018maximally} and \cite{lemieszAlgebraDataSketches2021}, respectively.
Therefore, we have the following error analysis results. 
\begin{thm}\cite{moulton2018maximally} 
When using the part $\vec{s}(\vec{v})$ of Gumbel-Max sketch to estimate the probability Jaccard similarity $\mathcal{J_P}(\vec{u},\vec{v})$ between $\vec{u}$ and $\vec{v}$,
the expectation and variance of estimation $\hat{\mathcal{J_P}}(\vec{u},\vec{v})$ are 
\[
\mathbb{E}\left(\hat{\mathcal{J_P}}(\vec{u},\vec{v})\right)= \mathcal{J_P}(\vec{u},\vec{v}),
\]
\[
\text{Var}\left(\hat{\mathcal{J_P}}(\vec{u},\vec{v})\right)= \frac{1}{k} \mathcal{J_P}(\vec{u},\vec{v})\left(1-\mathcal{J_P}(\vec{u},\vec{v})\right).
\]
\end{thm}
\begin{thm}\cite{lemieszAlgebraDataSketches2021}
When using the part $\vec{y}(\vec{v})$ of Gumbel-Max sketch to estimate the weighted cardinality $c_{\Pi}$ of a sequence $\Pi$,
the expectation and variance of estimation $\hat{c}_{\Pi}$ are 
\[
\mathbb{E}\left(\hat{c}_{\Pi}\right)= c_{\Pi},
\]
\[
\text{Var}\left(\hat{c}_{\Pi}/c_{\Pi}\right)= 2/k+\mathcal{O}(1/k^2) \approx 2/k.
\]
\end{thm}

\begin{algorithm}[t]
	\SetKwFunction{RandUNI}{RandUNI}
	\SetKwFunction{RandInt}{RandInt}
	\SetKwFunction{Return}{Return}
	\SetKwFunction{continue}{continue}
	\SetKwFunction{Swap}{Swap}
	\SetKwFunction{RandGamma}{RandGamma}
	\SetKwInOut{Input}{Input}
	\SetKwInOut{Output}{Output}
	\BlankLine
	
	\Input{$\vec{v} = (v_1, \ldots, v_n)$}
	\Output{$\vec{s} = (s_{1}, \ldots, s_{k})$, $\vec{y} = (y_{1}, \ldots, y_{k})$}
	\BlankLine
	$R\gets 0$;    $k^*\gets k$;
	$(y_1, \ldots, y_k) \gets (-1, \ldots, -1)$\;
	\ForEach {$i\in N^+_{\vec{v}}$}{
		$(b_i, z_i)\gets (0, 0)$; $(\pi_{i,1}, \ldots, \pi_{i,k})\gets (1, \ldots, k)$\;
	}
	\tcc{The following part is FastSearch}
	\While {$k^*\neq 0$}{
		$R\gets R + \Delta$\;
		\ForEach {$i\in N^+_{\vec{v}}$}{
			$R_i\gets \lceil R v_i^*\rceil$\;
			\While{$z_i < R_i$}{
					$z_i \gets z_i + 1$\;
					\tcc{Variable $u\sim \text{UNI}(0,1)$.}
					$u \gets \RandUNI(0, 1, seed \gets i||z_i)$\;
                	$b_i\gets b_i - \frac{1}{v_i}\left( \frac{\ln{u}}{k-z_i+1}\right) $\;\label{ln:ghash_ed}
                	\tcc{$\text{RandInt}(z_i, k)$ returns a number from $\{z_i,z_i+1,\ldots,k\}$ at random.}
                	$j\gets \RandInt(z_i, k)$\;\label{ln:pos_st}
                	\tcc{$\text{Swap}(\pi_{i, z_i}, \pi_{i, j})$ exchanges the values of two variables $\pi_{i, z_i}$ and $\pi_{i, j}$.}
                	$\Swap(\pi_{i, z_i}, \pi_{i, j})$\;\label{ln:pos_ed}
                	$c \gets \pi_{i, z_i}$\;

				\If{$y_c<0$}{ \label{ln:syj_st}
					$(y_c, s_c)\gets (b_i, i)$; 
					$k^*\gets k^* - 1$\;
				}
				\ElseIf{$b_i< y_c$}{
					$(y_c, s_c)\gets (b_i, i)$\; 
				}\label{ln:syj_ed}
			}
		}
	}
	\tcc{The following part is FastPrune}
	$j^* \gets \argmax_{j=1,\ldots,k} y_j$;
	$N\gets N^+_{\vec{v}}$\;
	\While {$N$ is not empty}{
		$R\gets R + \Delta$\;
		\ForEach {$i\in N$}{
			\While{$z_i < R_i$}{
				    $z_i \gets z_i + 1$\;
					$u \gets \RandUNI(0, 1, seed \gets i||z_i)$\;
                	$b_i\gets b_i - \frac{1}{v_i}\left( \frac{\ln{u}}{k-z_i+1}\right) $\;
                	$j\gets \RandInt(z_i, k)$\;
                	$\Swap(\pi_{i, z_i}, \pi_{i, j})$\;
                	$c \gets \pi_{i, z_i}$\;
				\If{$b_i > y_{j^*}$}{
					$N\gets N \setminus \{i\}$\;
					\textbf{break}\;
				}	
				\If{$b_i < y_c$}{
					$(y_c, s_c)\gets (b_i, i)$\; 
					\If {$c==j^*$}{
						$j^* \gets \argmax_{j=1,\ldots,k} y_j$\;
					}
				}
			}
		}
	}
	\caption{Pseudo code of our FastGM.\label{alg:FastGM-static}}
\end{algorithm}

\subsection{Space and Time Complexities}
\textbf{Space Complexity.} For a non-negative vector $\vec{v}$ with $n^+_{\vec{v}}$ positive elements,
our method FastGM requires $k\log k$ bits to store the $(\pi_{i,1}, \ldots, \pi_{i,k})$ of each $i\in N^+_{\vec{v}}$,
and in summary, $n^+_{\vec{v}} k\log k$ bits are desired.
In addition, $64k$ bits are desired for storing $y_1, \ldots, y_k$ (we use 64-bit floating-point registers to record $y_1, \ldots, y_k$), and $k\log n$ bits are required for storing $s_1, \ldots, s_k$, where $n$ is the size of the vector. 
However, the additional memory is released immediately after computing the sketch and is far smaller than the memory for storing the generated sketches of massive vectors (e.g. documents).
Therefore, FastGM requires $n^+_{\vec{v}} k\log k  +64k + k \log n$ bits when generating a $k$-length Gumbel-Max sketch $\vec{s}(\vec{v}) = (s_1, \ldots, s_k)$ and $\vec{y}(\vec{v}) = (y_1, \ldots, y_k)$ of $\vec{v}$.

\textbf{Time Complexity.}
We easily find that a non-negative vector and its normalized vector
have the same Gumbel-Max sketch.
For simplicity, therefore we analyze the time complexity of our method only for normalized vectors.
Let $\vec{v}^* = (v^*_1, \ldots, v^*_n)$ be a normalized and non-negative vector.
Define a variable $\tilde{y}^*$ as:
\[
\tilde{y}^* = \max_{j=1,\ldots,k} \tilde{y}_j,
\]
where
$
\tilde{y}_j = \min_{i\in N^+_{\vec{v}^*}} - \frac{\ln a_{i,j}}{v^*_i}$, $j=1, \ldots, k.
$
At the end of our FastPrune procedure, we easily find that each register $y_j$ used in the procedure equals $\tilde{y}_j$ and register $y^*$ equals $\tilde{y}^*$.
Because $-\frac{\ln a_{i,j}}{v^*_i} \sim \text{EXP}(v^*_i)$, we easily find that each $y_j$ follows the exponential distribution $\text{EXP}(\sum_{i=1}^{n} v^*_i)$, i.e. $\text{EXP}(1)$.
From~\cite{expectation}, we have
\[
	\mathbb{E}(\tilde y^*) = \sum_{m=1}^k \frac{1}{m}  \le  \ln k + \gamma, 
\]
\[
\text{Var}(\tilde y^*) = \sum_{m=1}^k \frac{1}{m^2} < \sum_{m=1}^\infty \frac{1}{m^2} = \frac{\pi^2}{6},
\]
where $\gamma = 1$.
From Chebyshev's inequality, we have
\[
P\left(|\tilde y^* - \mathbb{E}(\tilde y^*)| \ge \alpha \sqrt{\text{Var}(\tilde y^*)} \right) \le \frac{1}{\alpha^2}.
\]
Therefore, $\tilde y^*\le \mathbb{E}(\tilde y^*) + \alpha \sqrt{\text{Var}(\tilde y^*)}$ happens with a high probability when $\alpha$ is large. In other words, the random variable $\tilde y^*$ can be upper bounded by $\mathbb{E}(\tilde y^*) + \alpha \sqrt{\text{Var}(\tilde y^*)}$ with a high probability.
Next, we derive the expectation of $t_{i,R}$ after the first $R$ customers have been released.
For each queue $\mathcal{Q}_i$, $i\in N^+_{\vec{v}}$,
from Eqs.~(\ref{eq:xizexpvar}) and~(\ref{eq:exiri}), we find that
the last customer among these first $R$ customers has a timestamp $t_{i, R_i}$ with the expectation
$\mathbb{E}(t_{i, R_i}\mid R)  \approx \frac{R}{k}$.
When $R=k (\mathbb{E}(\tilde y^*) + \alpha \sqrt{\text{Var}(\tilde y^*))} < k(\ln k + \gamma + \frac{\alpha \pi}{\sqrt{6}})$, the probability of $\mathbb{E}(t_{i, R_i}) > \tilde y^*$ is almost 1 for large $\alpha$, e.g., $\alpha > 10$.
Therefore, we find that after the first $O(k\ln k)$ customers, each queue $\mathcal{Q}_i$ is expected to be early terminated and so we are likely to acquire all the Gumbel-Max variables.
We also note that each positive element has to be enumerated once in the FastPrune model. Therefore, the total time complexity of our method FastGM is $O(k \ln k+n^+_{\vec{v}})$.

\section{Our Method Stream-FastGM} \label{sec:extension}
We extend our method FastGM to handle data streams.
Given a stream $\Pi$ represented as a sequence of elements $i\in \{1, \ldots, n\}$.
An element $i$ may occur multiple times in $\Pi$ and it has a fixed weight $v_i$.
Our method Stream-FastGM is a fast one-pass algorithm for computing the Gumbel-Max sketch of $\Pi$, which reads and processes each element arriving at the stream exactly once.

The pseudo-code of Stream-FastGM is shown in Algorithm~\ref{alg:FastGM-stream}.
Similar to FastGM,  for each server $j=1, \ldots, k$,  we use two registers $y_j$ and $s_j$ to record its first customer's arrival time and queue number.
In addition, we use $y^* = \max_{j=1,\ldots,k} y_j$ to record the maximum of all $y_1, \ldots, y_k$.
As we mentioned, the FastPrune procedure can be used only after each of the servers has been selected by at least one customer.
We use a flag $FlagFastPrune$ to indicate whether the FastPrune procedure can be used.
For each element $i$ arriving at stream $\Pi$, we repeat to generate random exponential variables in ascending order. 
When the flag $FlagFastPrune$ is true and the generated variable has a value larger than $y^*$, we stop processing the current element. 

\begin{algorithm}[t]
	\SetKwFunction{RandUNI}{RandUNI}
	\SetKwFunction{RandInt}{RandInt}
	\SetKwFunction{Return}{Return}
	\SetKwFunction{continue}{continue}
	\SetKwFunction{Swap}{Swap}
	\SetKwFunction{RandGamma}{RandGamma}
	\SetKwInOut{Input}{Input}
	\SetKwInOut{Output}{Output}
	\BlankLine
	
	\Input{data stream $\Pi$}
	\Output{$\vec{s} = (s_{1}, \ldots, s_{k})$, $\vec{y} = (y_{1}, \ldots, y_{k})$}
	\BlankLine
	$k^*\gets k$;\quad $j^*\gets 1$; \quad
	$(y_1, \ldots, y_k) \gets (-1, \ldots, -1)$\;

	\ForEach {element $i$ in stream $\Pi$}{
		$b\gets 0$; $(\pi_{1}, \ldots, \pi_{k})\gets (1, \ldots, k)$\;
		\For {$l=1, \ldots, k$}{
					$u \gets \RandUNI(0, 1, seed \gets i||l)$\;
                	\tcc{$v_i$ is the weight of element $i$.}
                	$b\gets b - \frac{1}{v_i}\left( \frac{\ln{u}}{k-l+1}\right) $\;
                	$j\gets \RandInt(l, k)$\;
                	$\Swap(\pi_{l}, \pi_{j})$\;
                	$c \gets \pi_{l}$\;
				\If{FlagFastPrune==False}{
    				\If{$y_c<0$}{
    					$(y_c, s_c)\gets (b, i)$\; 
    					$k^*\gets k^* - 1$\;
                        \If{$k==0$}{
                        	$FlagFastPrune\gets True$\;
                        	$j^* \gets \argmax_{j=1,\ldots,k} y_j$\;
                        }
    				}
    				\ElseIf{$b< y_c$}{
    					$(y_c, s_c)\gets (b, i)$\; 
    				}
				}
				\If{FlagFastPrune==True}{
    				\If{$b > y_{j^*}$}{
    					\textbf{break}\;
    				}	
    				\If{$b < y_c$}{
    					$(y_c, s_c)\gets (b, i)$\; 
    					\If {$c==j^*$}{
    						$j^* \gets \argmax_{j=1,\ldots,k} y_j$\;
    					}
    				}
				}
		}
	}
	\caption{Pseudo code of our Stream-FastGM.\label{alg:FastGM-stream}}
\end{algorithm}

\section{Evaluation} \label{sec:results}
We evaluate our method FastGM with the state-of-the-art on two tasks: \textbf{(Task 1) probability Jaccard similarity estimation} and \textbf{(Task 2)  weighted cardinality estimation}.
All algorithms run on a computer with a Quad-Core Intel(R) Xeon(R) CPU E3-1226 v3 CPU 3.30GHz processor.
To demonstrate the reproducibility of the experimental results, we make our source code publicly available\footnote{https://github.com/YuanmingZhang05/FastGM}.

\subsection{Datasets}
For the task of probability Jaccard similarity estimation,
we verify the efficiency of our FastGM in generating Gumbel-Max sketch with different lengths $k\in\{2^6,2^7,\ldots,2^{12}\}$ for vectors of length in the range $n \in \{10^2, 10^3, 10^4\}$.
We generate the weights of synthetic vectors according to the uniform distribution $\text{UNI}(0,1)$ and the exponential distribution with rate 1 $\text{EXP}(1)$.
In addition,
we also run experiments on six real-world datasets: Real-sim~\cite{Wei2017Consistent}, Rcv$1$~\cite{Lewis2004RCV1}, News$20$~\cite{dataset:news20}, Libimseti~\cite{konect:2017:libimseti}, Wiki10~\cite{dataset:wiki10}, and MovieLens~\cite{konect:2017:movielens-10m_rating}.
In detail, Real-sim~\cite{Wei2017Consistent}, Rcv$1$~\cite{Lewis2004RCV1}, and 
News$20$~\cite{dataset:news20} are datasets of web documents from different sources 
where each vector represents a document and each entry in the vector refers to the TF-IDF score of a specific word for the document.
Libimseti~\cite{konect:2017:libimseti} is a dataset of ratings between users on the Czech dating site, where each vector refers to a user and each entry records the user's rating to another one.
Wiki$10$~\cite{dataset:wiki10} is a dataset of tagged Wikipedia articles, where each vector and element represent an article and a tag, respectively. 
Moreover, the weight of an element indicates how relevant the tag is for the article. 
MovieLens~\cite{konect:2017:movielens-10m_rating} is a dataset of movie ratings, where each vector is a user and each entry in the vector is that user's rating for a specific movie.
The statistics of all the above datasets are summarized in Table~\ref{tab:datasets}.

As for the task of weighted cardinality estimation,
we follow the experimental settings in~\cite{lemieszAlgebraDataSketches2021}.
We conduct experiments on both synthetic datasets and a simulated scenario obtained from real-world problems.
In the later Section~\ref{subsec:re_WCarEst}, we detail them.
In addition, we also design a data steaming setting to demonstrate the mergeability of our Gumbel-Max sketch and the performance of our Stream-FastGM.
Specifically, we generate a set of elements arriving in a streaming fashion.

\begin{table}[t]		
	\centering
	\caption{Statistics of used real-world datasets.}\label{tab:datasets}
		\begin{tabular}{|c|c|c|} \hline
			\textbf{Dataset} &\textbf{\#Vectors}&\textbf{\#Features}\\ \hline
			Real-sim \cite{Wei2017Consistent}&72,309&20,958\\ \hline
			Rcv$1$ \cite{Lewis2004RCV1}&20,242&47,236\\ \hline
			News$20$ \cite{dataset:news20}&19,996&1,355,191\\ \hline
			Libimseti \cite{konect:2017:libimseti}&220,970&220,970\\ \hline
			Wiki$10$ \cite{dataset:wiki10}&14,146&104,374\\ \hline
			MovieLens \cite{konect:2017:movielens-10m_rating}&69,878& 80,555 \\ \hline
		\end{tabular}
\end{table}
\subsection{Baseline}
To demonstrate the improvement of our FastGM over the conference version of FastGM (in short, \textbf{FastGM-c}),
we also apply FastGM-c as a baseline in efficiency experiments.
For task 1, \textbf{probability Jaccard similarity estimation}, we compare our method FastGM with $\mathcal{P}$-MinHash \cite{moulton2018maximally}.
To highlight the efficiency of FastGM, we further compare FastGM with the state-of-the-art weighted Jaccard similarity estimation method, BagMinHash~\cite{ertl2018bagminhash},
which is used for estimating weighted Jaccard similarity $\mathcal{J_W}$.
The weighted Jaccard similarity $\mathcal{J_W}$ that BagMinHash aims to estimate is an alternative similarity metric to the probability Jaccard similarity $\mathcal{J_P}$ we focused on in this paper.
Experiments and theoretical analysis in~\cite{moulton2018maximally} have shown that weighted Jaccard similarity $\mathcal{J_W}$ and probability Jaccard similarity $\mathcal{J_P}$ usually have similar performance on many applications such as fast searching similar set.
Notice that BagMinHash estimates a different similarity metric and thus we only show its results on efficiency.
For task 2, \textbf{weighted cardinality estimation}, we compare our method with Lemiesz's sketch~\cite{lemieszAlgebraDataSketches2021}.

\subsection{Metric}
For both tasks of probability Jaccard similarity estimation and weighted cardinality estimation,
we use the running time and root mean square error (RMSE) to measure our method's efficiency and effectiveness, respectively.
In detail, 
we measure the RMSEs of probability Jaccard similarity estimation $\hat{J}$ and weighted cardinality estimation $\hat{c}$
with respect to their true values $J$ and $c$ as:
\[
\text{RMSE}(\hat{J})=\sqrt{\mathbb{E}((\hat{J}-J)^2)},\quad \text{RMSE}(\hat{c})=\sqrt{\mathbb{E}((\hat{c}-c)^2)}.
\]
All experimental results are empirically computed from 1,000 independent runs by default.

\subsection{Probability Jaccard Similarity Estimation}
We conduct experiments on both synthetic and real-world datasets for the task of probability Jaccard similarity estimation. Specially, we first use synthetic weighted vectors to evaluate the performance of FastGM for vectors with different dimensions. Then, we show results on 
6 real-world datasets.

\begin{figure}[t]
\captionsetup[subfigure]{labelfont=bf}
\centering
    % \subfloat[$n = 10^{2}$]{\scalebox{0.48}{\input{fig9adata}}}
    % \subfloat[$n = 10^{3}$]{\scalebox{0.48}{\input{fig9bdata}}}\\
    % \subfloat[$n = 10^{4}$]{\scalebox{0.48}{\input{fig9edata}}}
    % \subfloat[$k = 2^8$]{\scalebox{0.48}{\input{fig9cdata}}}\\
    % \subfloat[$k = 2^9$]{\scalebox{0.48}{\input{fig9ddata}}}
    % \subfloat[$k = 2^{10}$]{\scalebox{0.48}{\input{fig9fdata}}}\\
    \subfloat[$n = 10^{2}$]{\scalebox{0.48}{\includegraphics{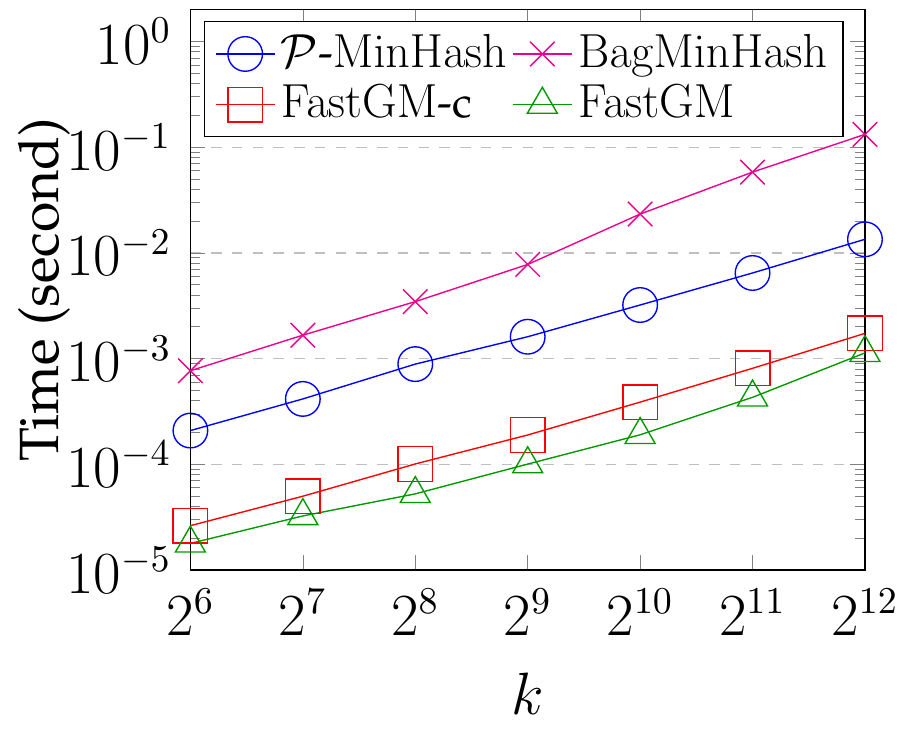}}}
    \subfloat[$n = 10^{3}$]{\scalebox{0.48}{\includegraphics{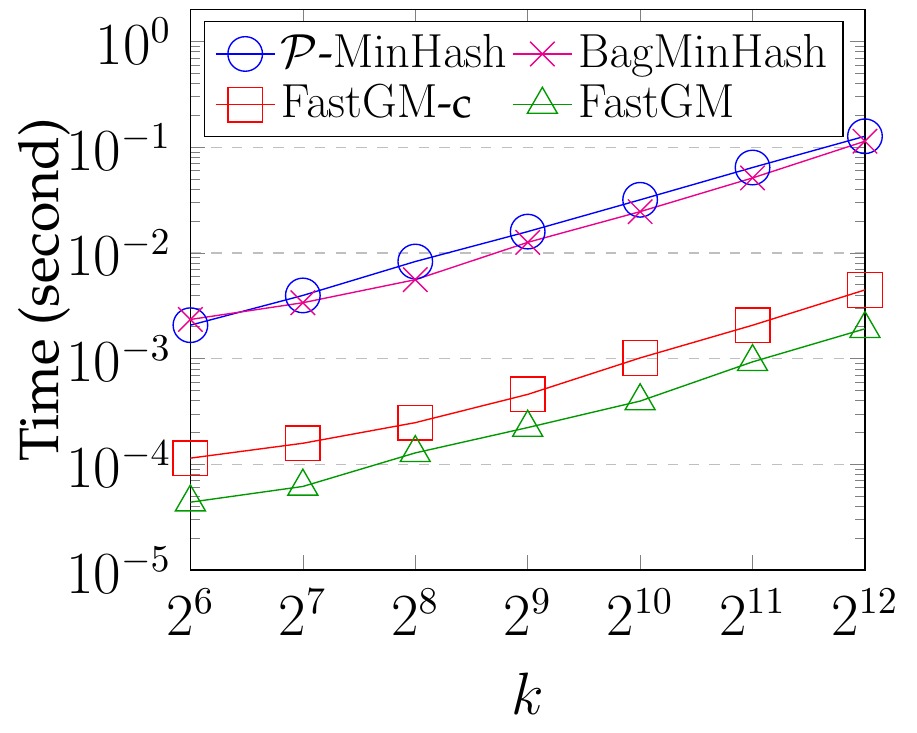}}}\\
    \subfloat[$n = 10^{4}$]{\scalebox{0.48}{\includegraphics{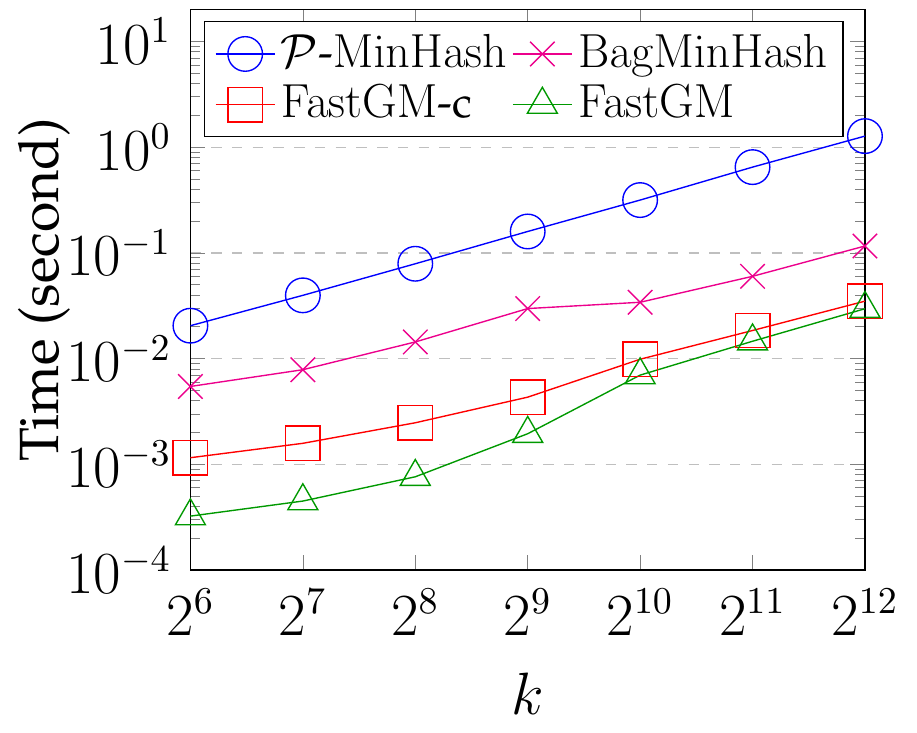}}}
    \subfloat[$k = 2^8$]{\scalebox{0.48}{\includegraphics{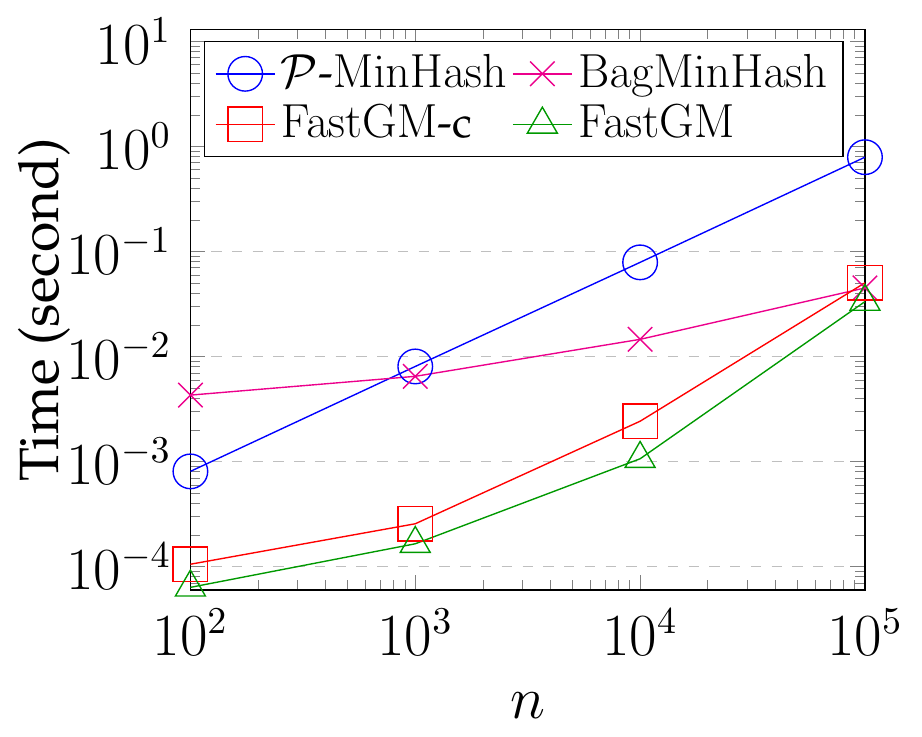}}}\\
    \subfloat[$k = 2^9$]{\scalebox{0.48}{\includegraphics{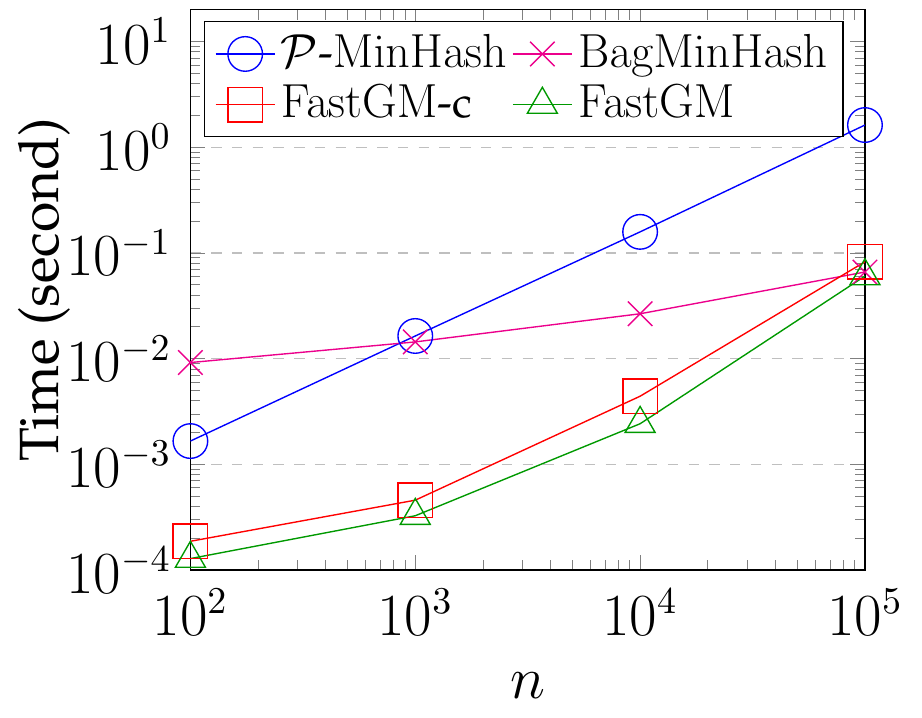}}}
    \subfloat[$k = 2^{10}$]{\scalebox{0.48}{\includegraphics{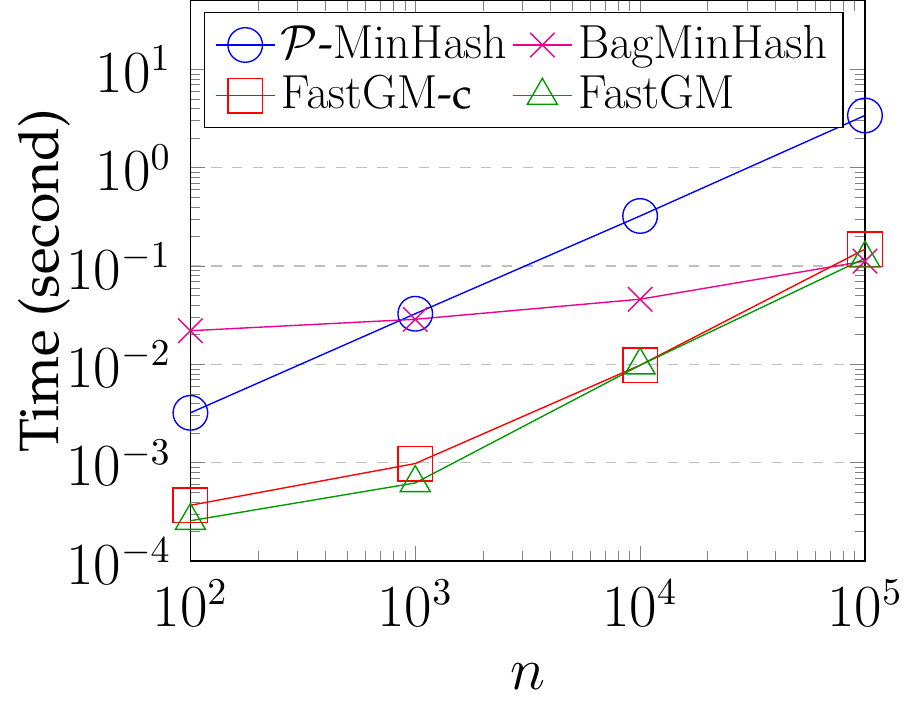}}}\\
    \caption{(Task 1) The efficiency of FastGM compared with $\mathcal{P}$-MinHash, BagMinHash, and the conference version of FastGM (FastGM-c) on synthetic vectors, where each element in the vector is randomly selected from UNI(0,1).}\label{fig:syndataset}
\end{figure}

\textbf{Results on synthetic vectors.}
We first conduct experiments on weighted vectors with uniform-distribution weights.
Without loss of generality, we let $n^+_{\vec{v}} = n$ for each vector, i.e., all elements of each vector are positive.
As shown in Fig.~\ref{fig:syndataset} (a), (b) and (c), when $n=10^3$, FastGM is $13$ and $22$ times faster than BagMinHash and $\mathcal{P}$-MinHash respectively.
As $n$ increases to $10^4$, the improvement becomes $8$ and $125$ times respectively.
Especially, the sketching time of our method is around $0.02$ seconds when $n=10^4$ and $k=2^{12}$, while BagMinHash and $\mathcal{P}$-MinHash take over $0.15$ and $2.5$ seconds for sketching respectively.
In Fig.~\ref{fig:syndataset} (d), (e), and (f), we show the running time of all competitors for different $n$.
Our method FastGM is $13$ to $100$ times faster than $\mathcal{P}$-MinHash for different $n$.
Compared with BagMinHash, FastGM is about $60$ times faster when $n=1,000$,
and is comparable as $n$ increases to $100,000$.
It indicates that our method FastGM significantly outperforms BagMinHash
for vectors having less than  $100,000$ positive elements,
which are prevalent in real-world datasets.
As shown in Fig.~\ref{fig:syndataset}, our FastGM is consistently faster than FastGM-c, when $n=100$ and $n=1,000$ FastGM is around $1.2$ and $1.5$ times faster than FastGM-c, respectively.
Results are similar when the weights of synthetic vectors follow the exponential distribution $\text{EXP}(1)$,
thus we omit them here.

\begin{figure}[t]
\captionsetup[subfigure]{labelfont=bf}
\centering
    \subfloat[Real-sim]{\scalebox{0.48}{\includegraphics{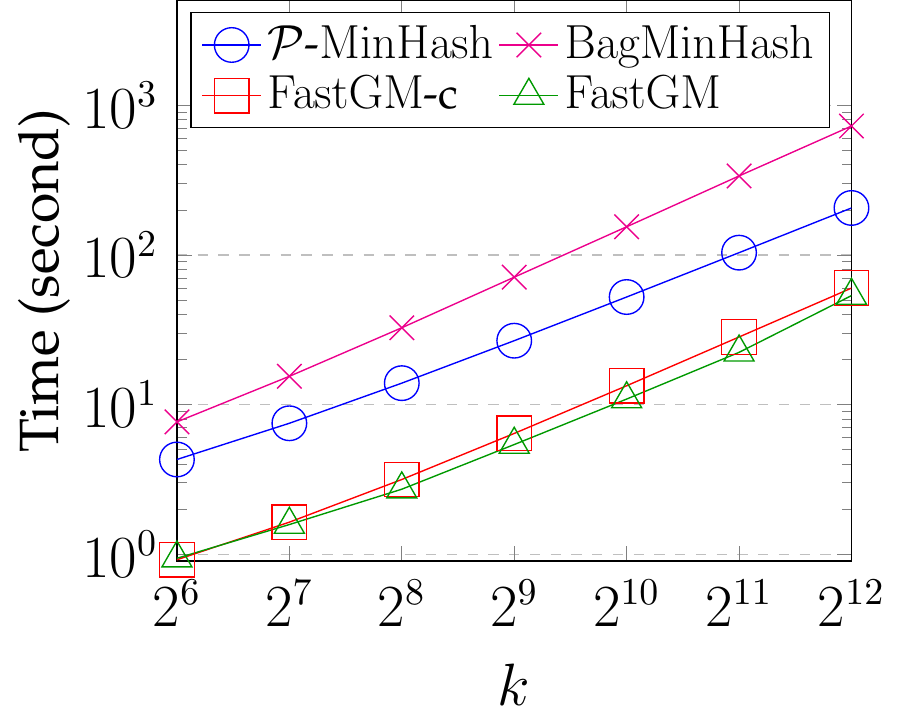}}}
    \subfloat[Rcv1]{\scalebox{0.48}{\includegraphics{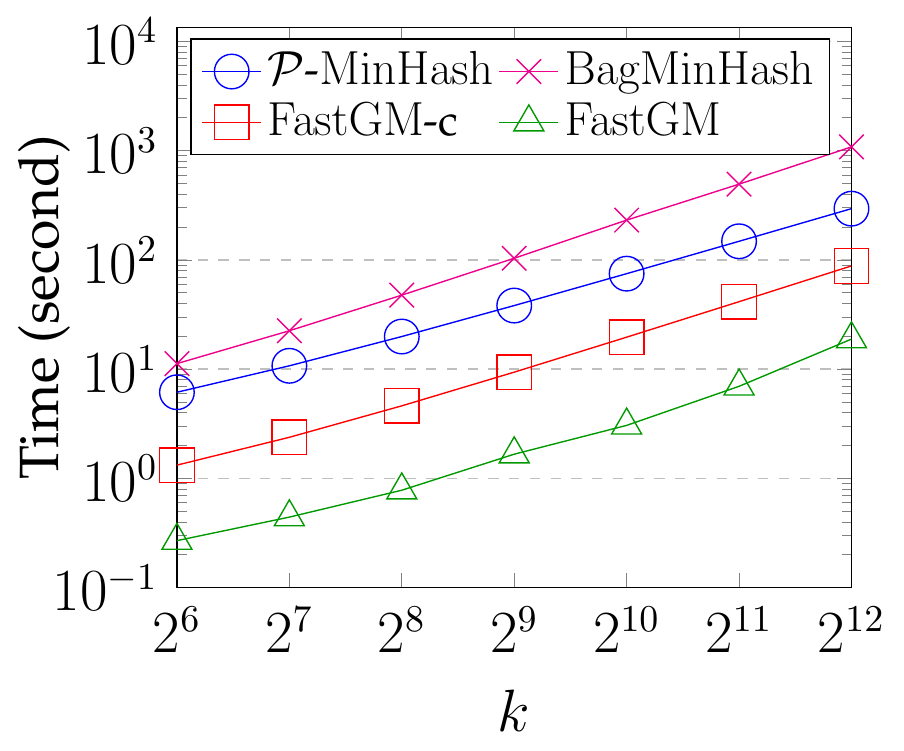}}}\\
    \subfloat[News20]{\scalebox{0.48}{\includegraphics{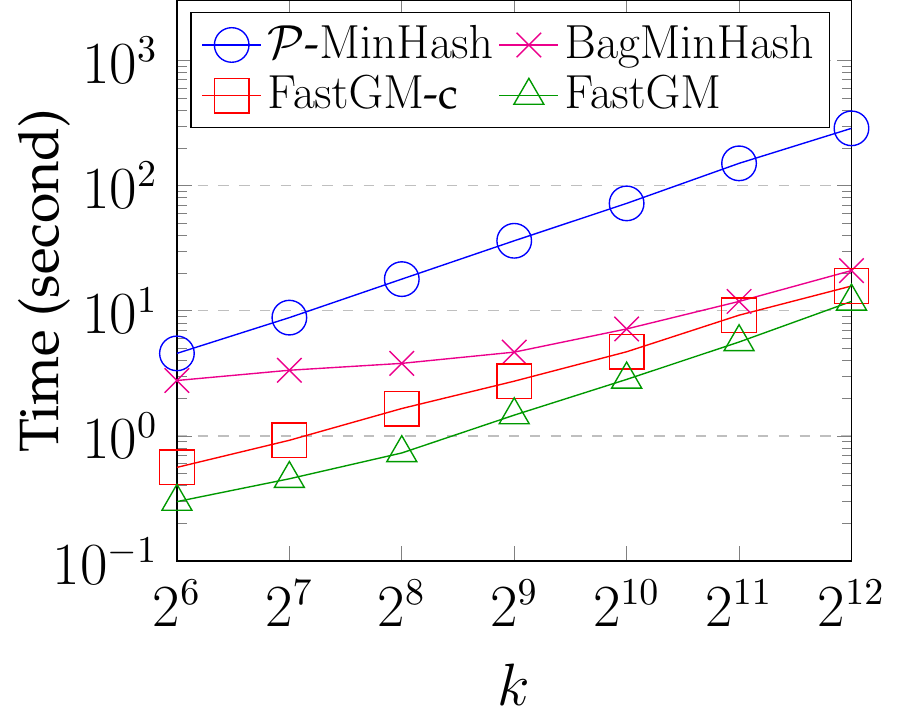}}}
    \subfloat[Libimseti]{\scalebox{0.48}{\includegraphics{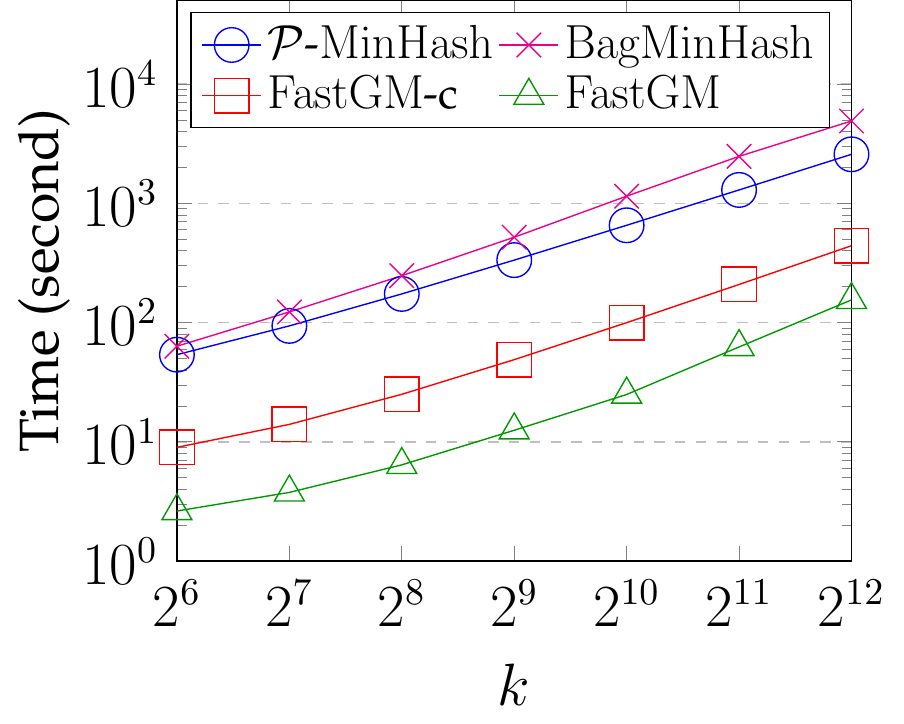}}}\\
    \subfloat[Wiki10]{\scalebox{0.48}{\includegraphics{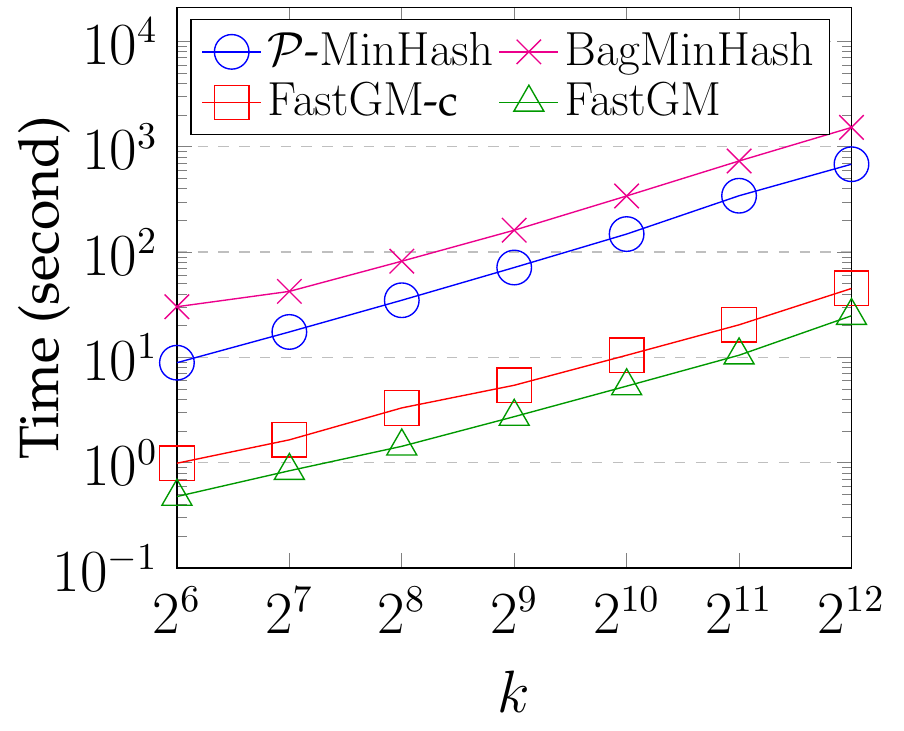}}}
    \subfloat[MovieLens]{\scalebox{0.48}{\includegraphics{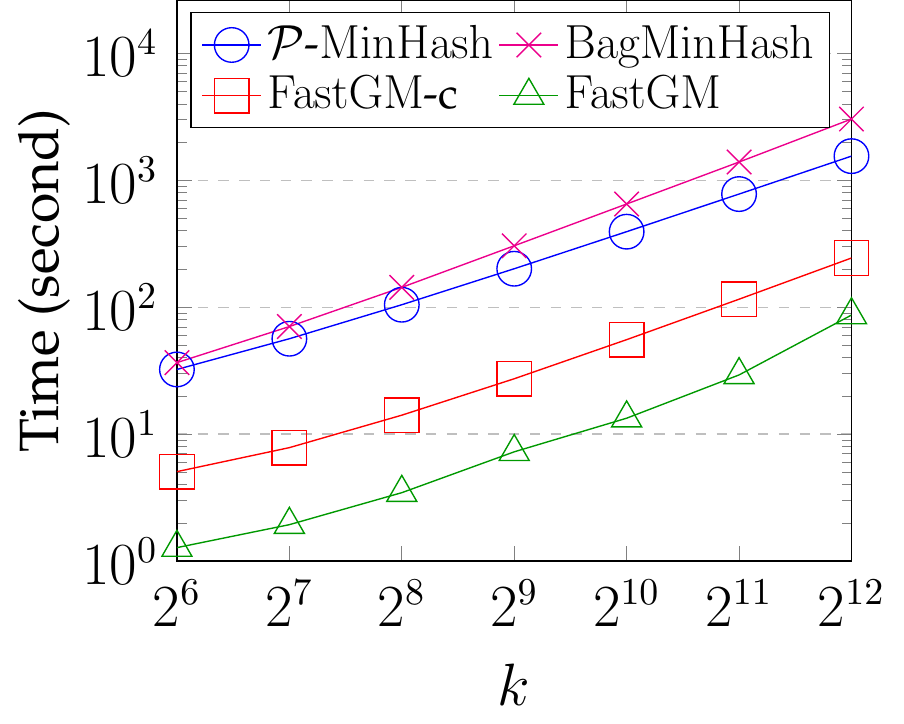}}}\\
    \caption{(Task 1) Efficiency of FastGM compared with $\mathcal{P}$-MinHash, BagMinHash, and the conference version of FastGM (FastGM-c) for different $k$ on real-world datasets.}\label{fig:time_realdata}
\end{figure}

\begin{figure}[t]
\captionsetup[subfigure]{labelfont=bf}
\centering
    \subfloat[Real-sim]{\scalebox{0.48}{\includegraphics{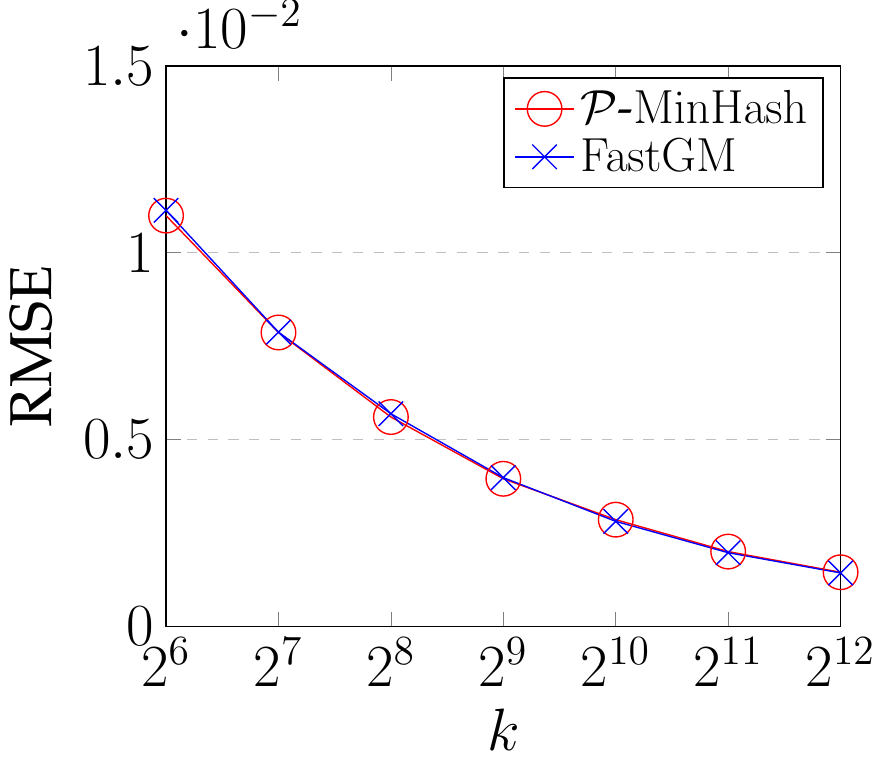}}}
    \subfloat[MovieLens]{\scalebox{0.48}{\includegraphics{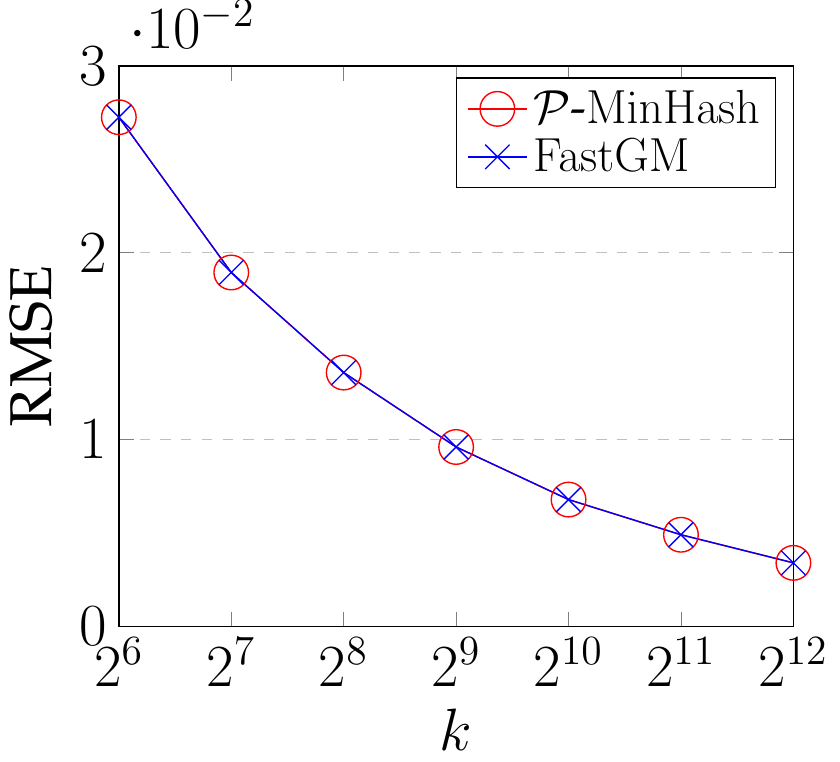}}}\\
    \caption{(Task 1) The accuracy of FastGM compared with $\mathcal{P}$-MinHash for different $k$.}\label{fig:rmse}
\end{figure}

\textbf{Results on real-world datasets.}
Next, we show results on the real-world datasets in Table~\ref{tab:datasets}.
We report the sketching time of all algorithms in Fig.~\ref{fig:time_realdata}.
We see that our method outperforms $\mathcal{P}$-MinHash and BagMinHash on all the datasets.
FastGM is consistently faster than FastGM-c, especially on datasets Rcv$1$, Libimesti, and MovieLens, FastGM is $4$ times faster than FastGM-c on average.
On sparse datasets such as Real-sim, Rcv$1$, Wiki$10$, and MovieLens, FastGM is about $8$ and $12$ times faster than $\mathcal{P}$-MinHash and BagMinHash respectively.
BagMinHash is even slower than $\mathcal{P}$-MinHash on these datasets.
On dataset News$20$ we note that FastGM is $26$ times faster than $\mathcal{P}$-MinHash.

Fig.~\ref{fig:rmse} shows the estimation error of FastGM and $\mathcal{P}$-MinHash on datasets Real-sim and MovieLens.
Due to a large number of vector pairs, we here randomly select $100,000$ pairs of vectors from each dataset and report the average RMSE.
We note that both algorithms give similar accuracy, which is coincident with our analysis.
We omit similar results on other datasets.

\subsection{Weighted Cardinality Estimation}\label{subsec:re_WCarEst}
In this task, we compare our FastGM with Lemiesz's sketch on both effectiveness and efficiency. The experimental results show that our FastGM sketch has the same accuracy as Lemiesz's sketch and is orders of magnitude faster in producing sketches. In the following, we detail the experiments on both synthetic datasets and a simulated scenario obtained from real-world problems in wireless sensor networks.

\begin{figure}[t]
\captionsetup[subfigure]{labelfont=bf}
\centering
    \subfloat[$v_i\sim$ \text{UNI}$(0, 1)$, $k=200$]{\scalebox{0.45}{\includegraphics{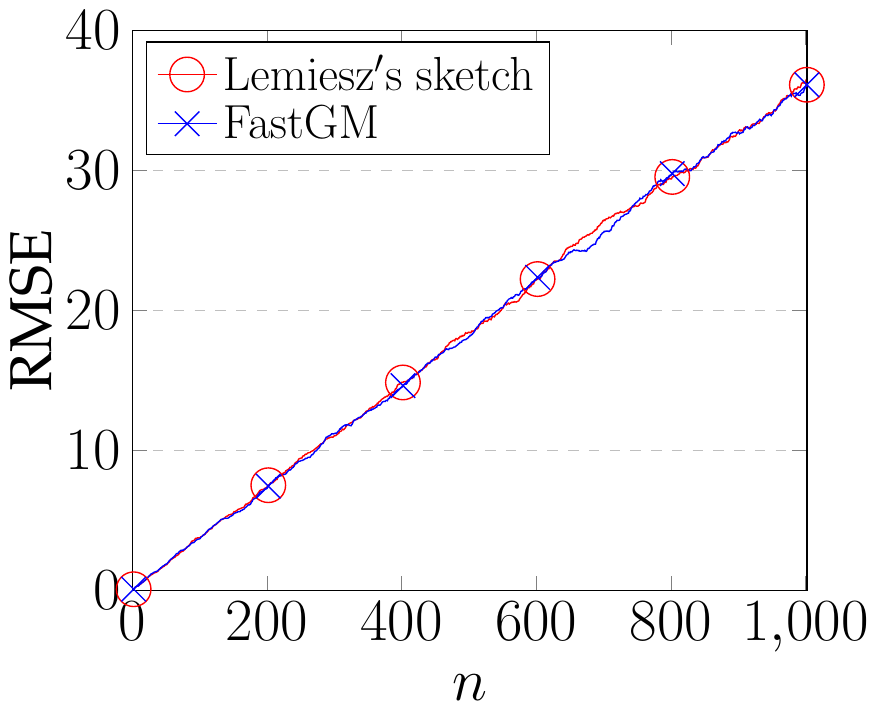}}} \quad
    \subfloat[$v_i\sim N(1, 0.1)$, $k=200$]{\scalebox{0.45}{\includegraphics{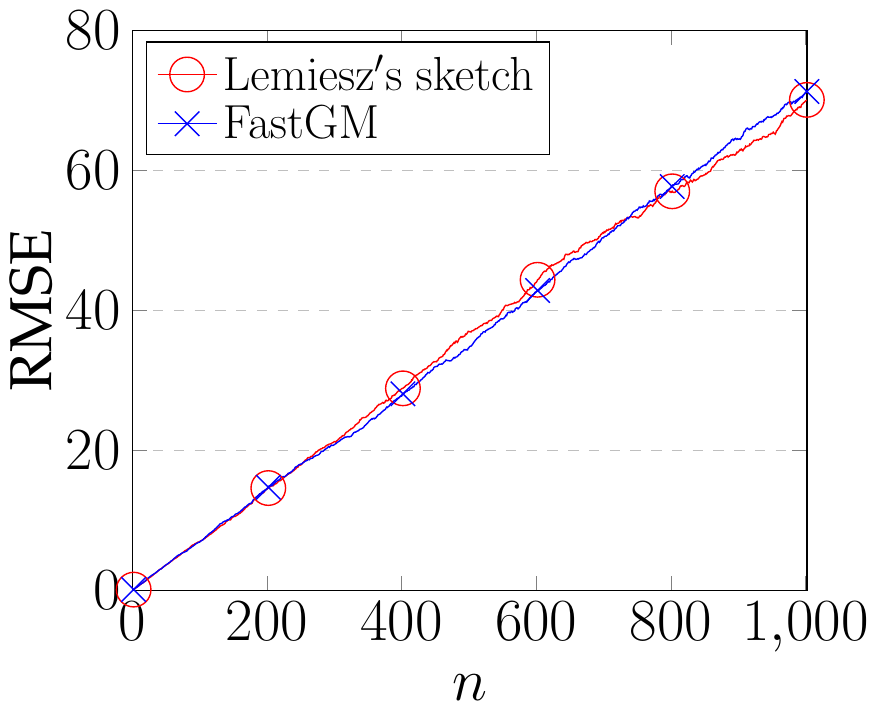}}}\\
    \subfloat[$v_i\!\sim$ \!\text{UNI}$(0, 1)$, $n=1,000$]{\scalebox{0.45}{\includegraphics{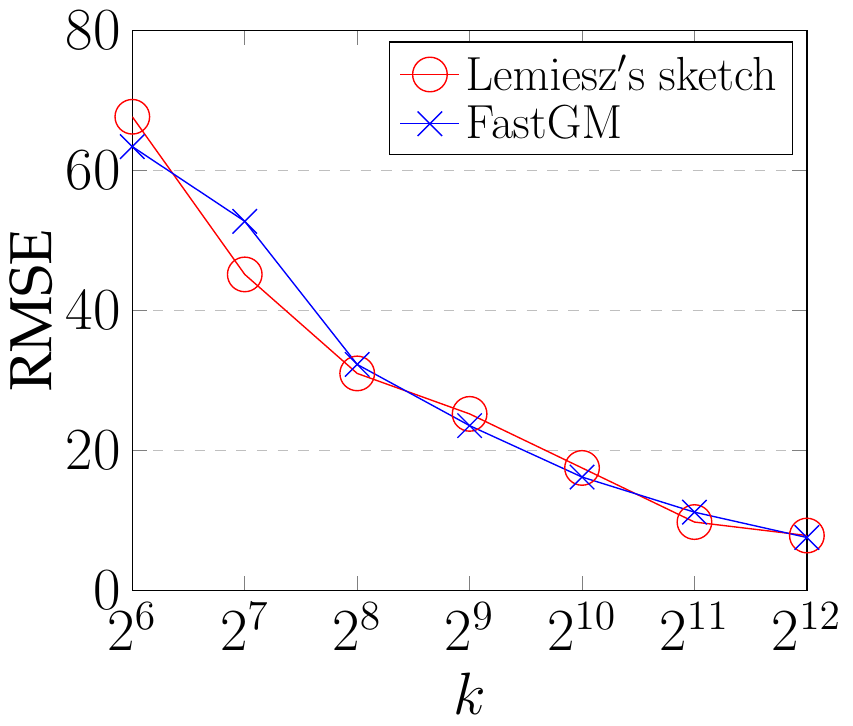}}} \quad
    \subfloat[$v_i\!\sim\! N(1, 0.1)$, $n=1,000$]{\scalebox{0.45}{\includegraphics{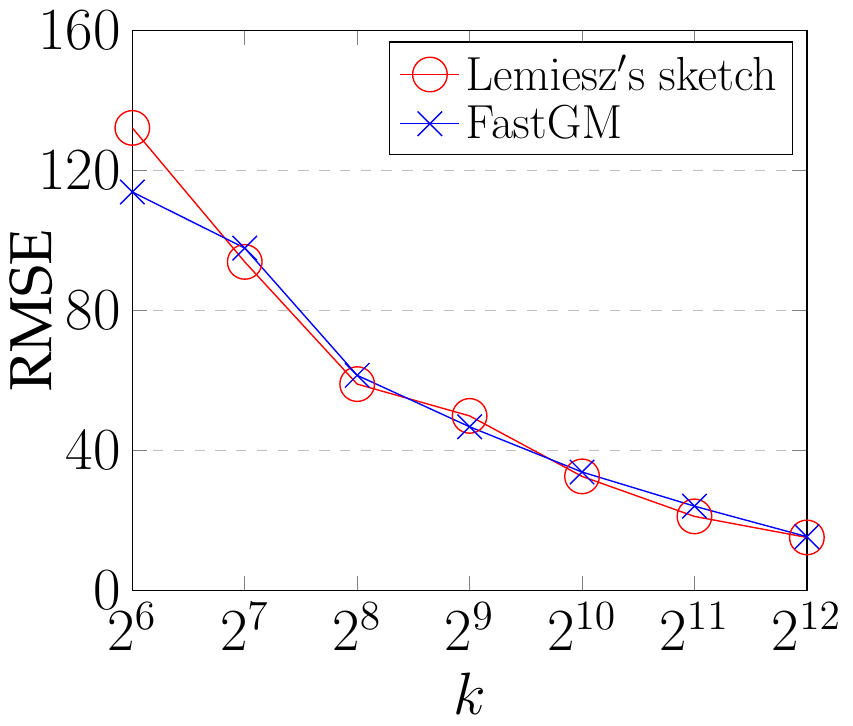}}}
    \caption{(Task 2) The weighted cardinality estimation errors on synthetic datasets, where the lengths $k$ of both Lemiesz's and FastGM sketches are the same.
    For each data example,
    we generate $\vec{v}^{\Pi}$ with $n$ objects of which weights $\vec{v}^{\Pi}_j$ are derived  according to the uniform distribution \text{UNI}$(0,1)$ and the normal distribution $N(1,0.1)$ respectively.}\label{fig:t2_effect}
\end{figure}

\begin{figure}[t]
\captionsetup[subfigure]{labelfont=bf}
\centering
    \subfloat[Generating sketches of length $k$ for data examples with $n=1,000$
    objects\label{fig:t2_syn_efc}]{\scalebox{0.45}{\includegraphics{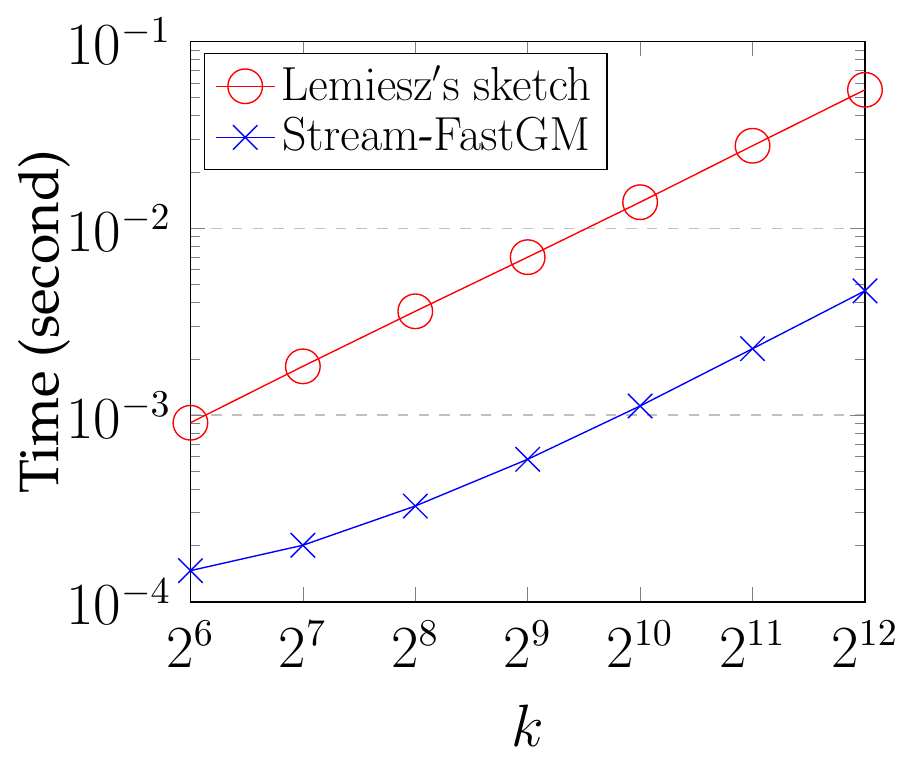}}}\;
    \subfloat[Generating $k=1024$ length sketches for data examples with $n$ objects
    \label{fig:t2_syn_efc_2}]{\scalebox{0.46}{\includegraphics{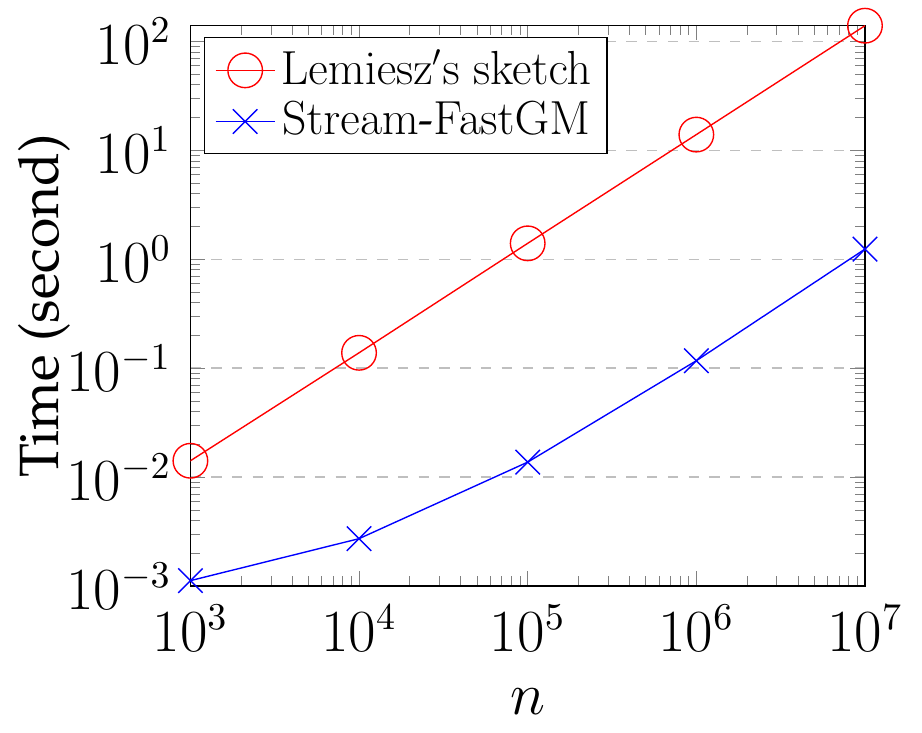}}}
    \caption{(Task 2) Average running time of Stream-FastGM and Lemiesz's sketch on synthetic data.}
\end{figure}

\textbf{Results on synthetic datasets.}
To evaluate the weighted cardinality estimation accuracy of our method,  
we generate a variety of data examples with different cardinalities.
We vary the number of elements in the data examples and generate the weights of elements according to the uniform distribution \text{UNI}$(0,1)$ and the normal distribution $N(1,0.1)$.
We report the RMSEs between the true cardinalities $c$ of data examples and estimations $\hat{c}$ from the sketches.
As shown in Fig.~\ref{fig:t2_effect}, our FastGM sketch has the same performance as Lemiesz's sketch on each dataset,
because the $\vec{y}$ part of FastGM and Lemiesz's sketch have the same results but are computed in different ways.
The efficiency of generating the two sketches is totally the same as the results reported in Fig.~\ref{fig:syndataset}, 
where Lemiesz's sketch has the same running time as $\mathcal{P}$-MinHash.
Therefore, in terms of efficiency, our FastGM sketch outperforms Lemiesz's sketch
by as much as FastGM outperforms $\mathcal{P}$-MinHash.
Hence we omit similar results.
Moreover,
in Fig.~\ref{fig:t2_syn_efc} we show the running time of computing the sketches by using our
Stream-FastGM compared with Lemiesz's sketch, and our Stream-FastGM is $23$ times faster than Lemiesz's sketch on average when $n=1,000$.
In Fig.~\ref{fig:t2_syn_efc_2} we report the running time of generating the sketches of length  $k=1024$ for 
data examples with different objects $n$, our Stream-FastGM is about $120$ times faster than Lemiesz's sketch at $n=10^6$.

\begin{figure}[t]
	\centering
	\includegraphics{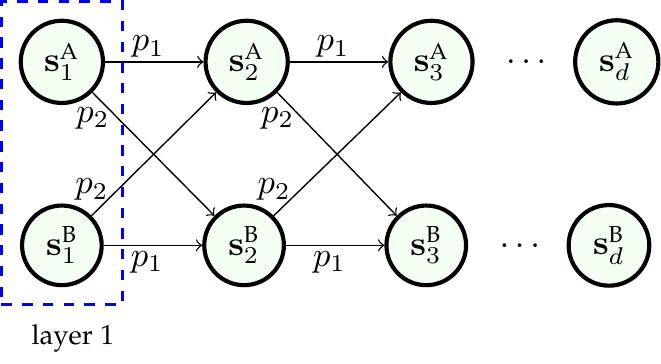}
	\caption{An example to show a simulated sensor network with $d$ layers using the braid chain strategy to transfer traffic packets, 
	where each node and each edge represent a sensor in the network and a traffic transfer path, respectively. The $p_1$ and $p_2$ denote the probability of a successful transmission between two nodes.
	}
	\label{fig:BCnetwork}
\end{figure}

\textbf{Results on the simulated scenario.}
Following the experimental setting in~\cite{lemieszAlgebraDataSketches2021}, 
we conduct experiments on simulated multi-hop wireless sensor networks 
where sensors use a braid chain strategy to guarantee the robustness of communication. 
In Fig.~\ref{fig:BCnetwork}, we show the topology of simulated networks.  
A braided chain consists of two sequences of nodes (sensors) 
$\mathbb{S}^\text{A}=[\mathbf{s}^\text{A}_1,\cdots,\mathbf{s}^\text{A}_d]$ and $\mathbb{S}^\text{B}=[\mathbf{s}^\text{B}_1,\cdots,\mathbf{s}^\text{B}_d]$.
Nodes with the same position in the sequences, such as $\mathbf{s}^\text{A}_1$ and $\mathbf{s}^\text{B}_1$, 
are considered as nodes in the same layer. 
Because the transfer path (edges in the network topology) is unstable.
To guarantee the transfer of traffic packets, 
the node in the previous layer \emph{\textbf{redundantly}} transfers traffic packets to all nodes in the next layer.
Specifically, the transfer path between nodes in the same sensors sequence and between different sensors sequences
(e.g., the edge between $\mathbf{s}^\text{A}_1$ and $\mathbf{s}^\text{A}_2$, the edge between $\mathbf{s}^\text{A}_1$ and $\mathbf{s}^\text{B}_2$) work well in chance $p_1$ and $p_2$, respectively.
For example, a traffic packet in node $\mathbf{s}^\text{A}_1$ is successfully sent to  $\mathbf{s}^\text{A}_2$ in chance $p_1$,
meanwhile a copy of this traffic packet also has $p_2$ chance to be successfully sent to node $\mathbf{s}^\text{B}_2$.
Note that $p_1$ does not necessarily equal $1-p_2$.

In the experiment setting,
the first node in each sequence is considered as the source that generates traffic packet $i$ with size $v_i$ in sequence.
After each source $\mathbf{s}_1$ generates a sequence $\Pi$ consisting of $n$ traffic packets, 
we have a vector $\vec{v}^{\Pi}$ of length $n$ from this sequence $\Pi$.
In our experiment, we follow the setting in \cite{lemieszAlgebraDataSketches2021} 
and set $p_1=0.9$, $p_2=0.1$, $d=30$, $n=10,000$ 
and the sizes of packets $v_i$ are generated according to a Beta distribution with parameters $\alpha=\beta=5$.
Take a node in the second layer $\mathbf{s}^\text{A}_2$ as an example, 
traffic packet sequence received by $\mathbf{s}^\text{A}_2$ is a mixture of some traffic packets in  
sequences $\Pi_{\scaleto{\mathbf{s}^\text{A}_1}{10pt}}$ and $\Pi_{\scaleto{\mathbf{s}^\text{B}_1}{10pt}}$ of both sources $\mathbf{s}^\text{A}_1$ and $\mathbf{s}^\text{B}_1$.
For the traffic packet sequence that passes through each node in the network, we build a sketch for it and use the sketch to estimate the total size of \textbf{\emph{distinct}} packets appearing in this sequence.
In this case, the weighted cardinality of the sequence represents the sum of distinct packets' sizes in the sequence. 
The reasons to build a sketch rather than simply use a counter are: 1) the traffic packet sequences passing through nodes in layers behind the second layer contain repetitive traffic packets, which causes the double-counting problem;
2) aggregating sketches rather than packets will not cause an explosion of packets even when the network follows a flooding strategy while guaranteeing the certain accuracy of network communication~\cite{lemieszAlgebraDataSketches2021}.
More than that, based on the sketches we are able to obtain more useful information about the network and we conduct the following experiments to demonstrate this. Ground truth results are shown as solid lines, and estimations obtained from sketches are shown as dashed lines. \emph{For clarity, we use the symbol $N_{\scaleto{\mathbf{s}}{4pt}}$ to represent the weighted set of packets that occurred in traffic packet sequence $\Pi_{\scaleto{\mathbf{s}}{4pt}}$ of a node rather than $N_{\Pi_{\scaleto{\mathbf{s}}{3pt}}}$.}
\begin{figure}[t]
\centering
\captionsetup[subfigure]{labelfont=bf}
    \subfloat[labelfont=bf][
    The total size of distinct packets from sources A and B at node $\mathbf{s}^{\text{A}}_\ell$ in each layer.
    \label{fig:re_sim1}]{\scalebox{0.49}{\includegraphics{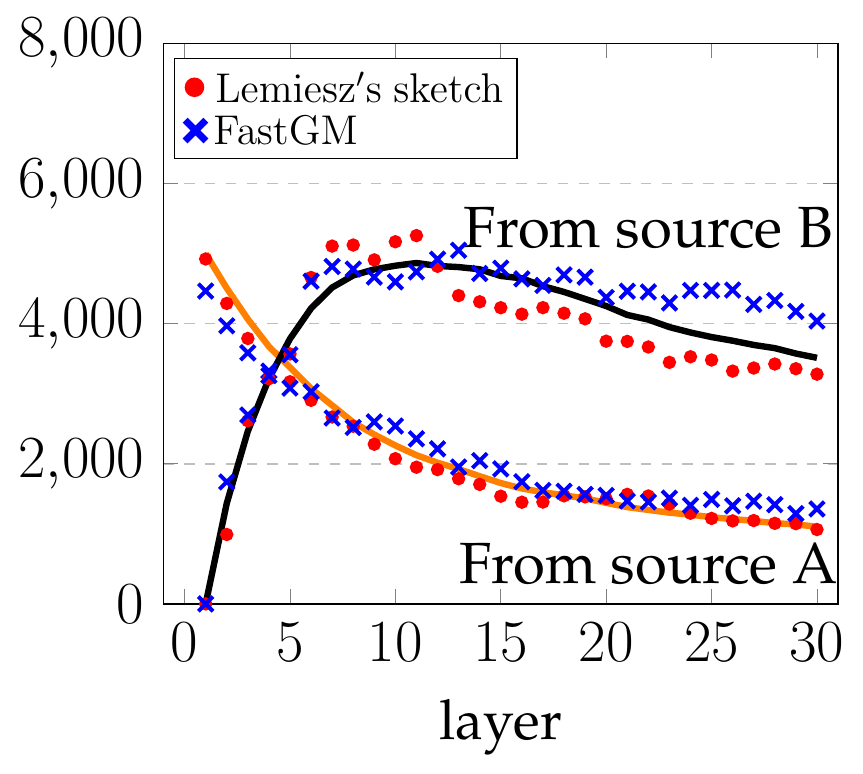}}}
    \quad
    \subfloat[The mean size of distinct packets passing through node $\mathbf{s}^{\text{A}}_\ell$ in each layer.
    \label{fig:re_sim2}]{\scalebox{0.49}{\includegraphics{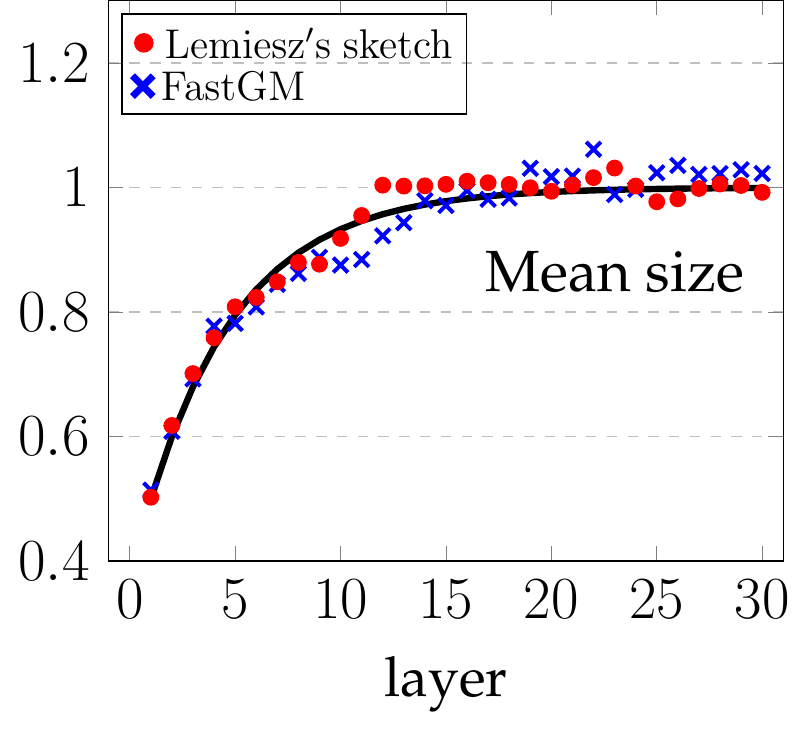}}}\\
    \subfloat[The total size of lost packets sent from source $\mathbf{s}^{\text{A}}_1$ in each layer. \label{fig:re_sim3}]{\scalebox{0.49}{\includegraphics{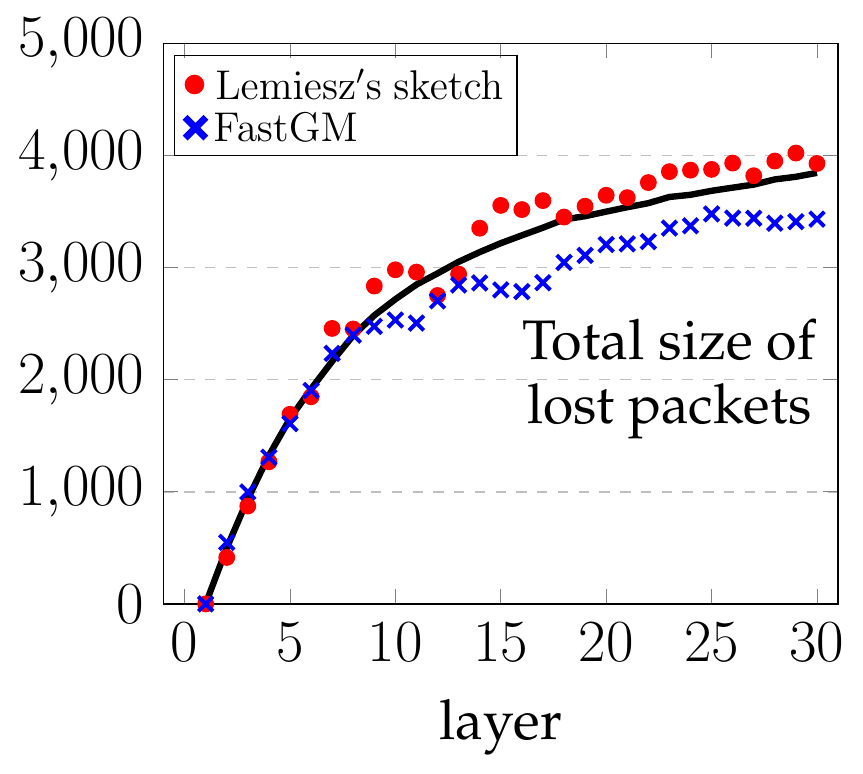}}}
    \quad
    \subfloat[The weighted Jaccard similarity between two nodes in each layer.\label{fig:re_sim4}]{\scalebox{0.49}{\includegraphics{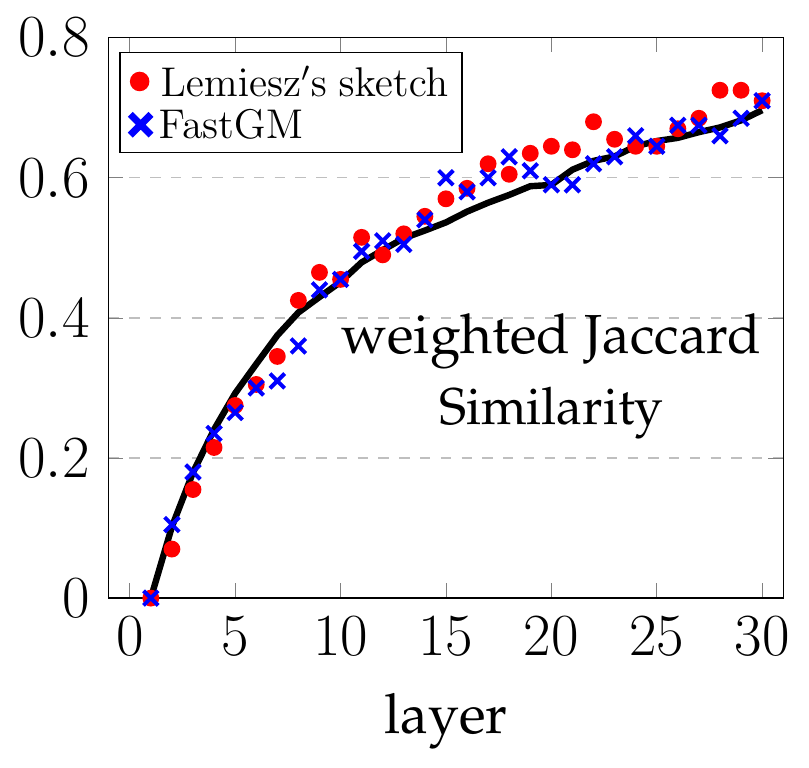}}}
    \caption{(Task 2) In Fig.~\ref{fig:re_sim1}-\ref{fig:re_sim4} solid lines represent ground truth results, 
     red circle and blue cross points respectively represent the estimation results based on Lemiesz's sketches and FastGM sketches with length $k = 200$. All results are obtained from a simulated sensor network with $d=30$ layers where each data source generates $n=10,000$ packets.}
\end{figure}

\begin{figure}[t]
\captionsetup[subfigure]{labelfont=bf}
\centering
    \subfloat[Generating sketches of different lengths $k$ on each node of simulated sensor networks with $d=30$ layers.\label{fig:t2_simu_efc}]{\scalebox{0.45}{\includegraphics{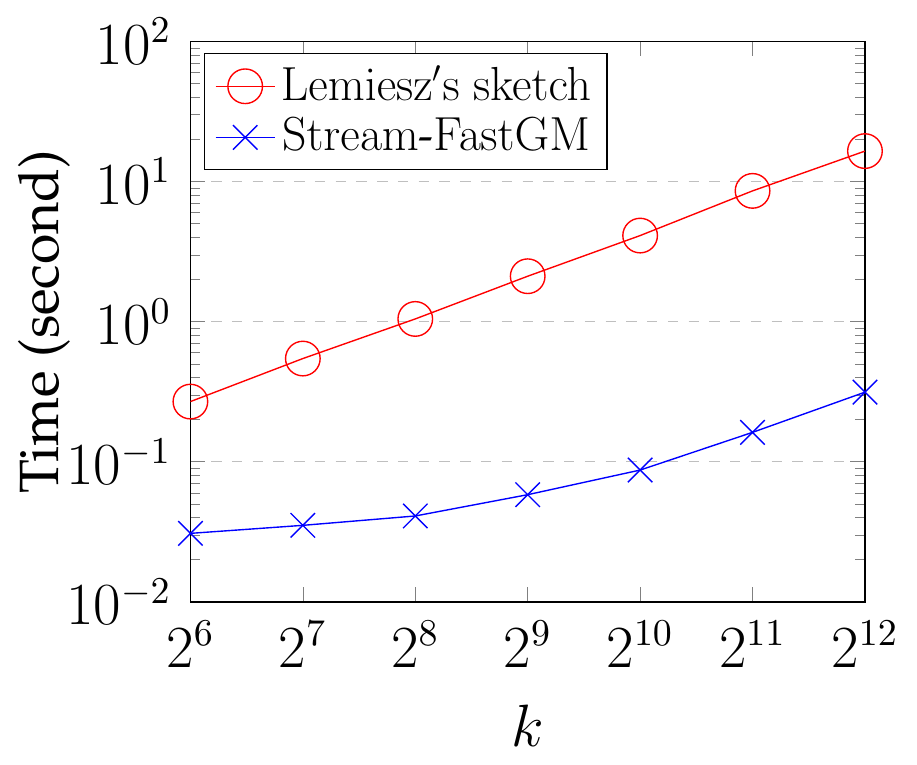}}} \quad
    \subfloat[Generating sketches of length $k=1024$ for nodes of simulated networks with different
    depth of layers.\label{fig:t2_simu_efc_2}]{\scalebox{0.45}{\includegraphics{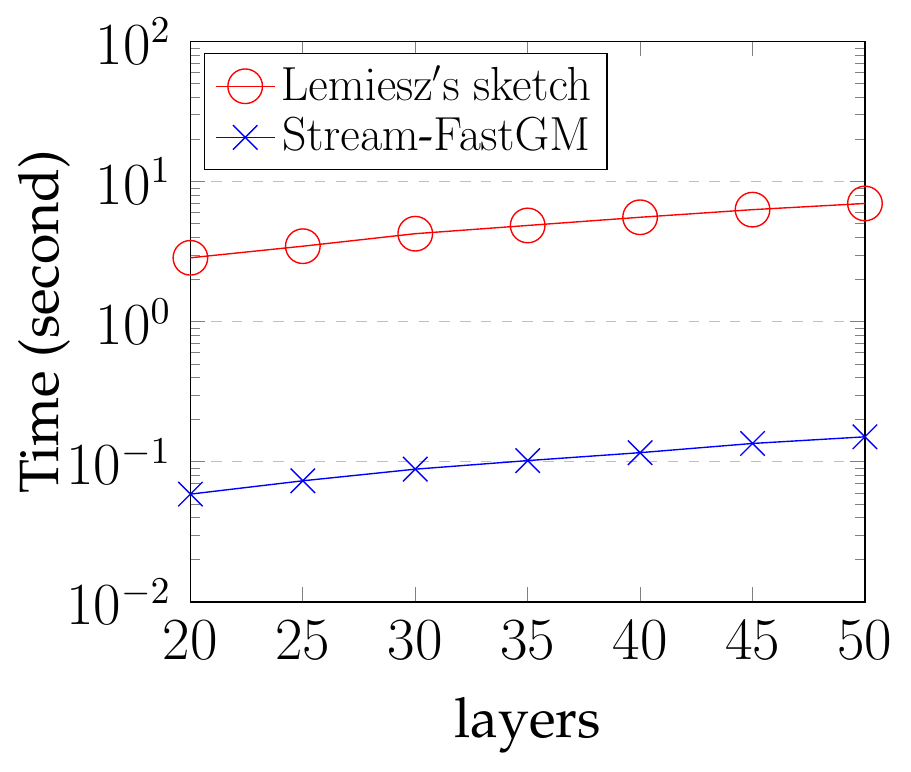}}}
    \caption{(Task 2) Average running time of Stream-FastGM and Lemiesz's sketch on simulated sensor networks.}
\end{figure}
In Fig.~\ref{fig:re_sim1}, black and orange lines represent the size of packets from source $\mathbf{s}^\text{A}_1$ and $\mathbf{s}^\text{B}_1$ respectively.
The size of distinct packets received by a node $\mathbf{s}^\text{A}_\ell$, $1\leq \ell \leq d$ is defined as
$
|N_{\scaleto{\mathbf{s}^\text{A}_\ell}{10pt}}|_\text{w}=
\sum_{\scaleto{i\in N}{7.5pt}{\scaleto{\mathbf{s}^\text{A}_\ell}{9pt}}} v_i,
$
where $i$ represents a traffic packet and $v_i$ is the size of packet $i$.
For a  node $\mathbf{s}^\text{A}_\ell$ in sequence $\mathbb{S}^A$, the sizes of distinct packets sent from source $\mathbf{s}^\text{A}_1$  and source $\mathbf{s}^\text{B}_1$ are computed as
$
|N_{\scaleto{\mathbf{s}^\text{A}_1}{10pt}}\cap N_{\scaleto{\mathbf{s}^\text{A}_\ell}{10pt}}|_\text{w}
$ and
$
|N_{\scaleto{\mathbf{s}^\text{B}_1}{10pt}}\cap N_{\scaleto{\mathbf{s}^\text{A}_\ell}{10pt}}|_\text{w}
$, respectively.
In Fig.~\ref{fig:re_sim2}, we show the results of estimating the average size of distinct packets on each node in sensor sequence $\mathbb{S}^A$.
In Fig.~\ref{fig:re_sim3}, we use the sketches to estimate the total size of lost packets from source $\mathbf{s}_1^\text{A}$ in each layer of the braided chain.  
The set of lost packets from source $\mathbf{s}_1^\text{A}$ in a layer $N_{\scaleto{\mathbf{L}^\text{A}_\ell}{10pt}}$ can be obtained from 
$N_{\scaleto{\mathbf{L}^\text{A}_\ell}{10pt}} = N_{\scaleto{\mathbf{s}^\text{A}_1}{10pt}}\setminus (N_{\scaleto{\mathbf{s}^\text{A}_\ell}{10pt}} \cup N_{\scaleto{\mathbf{s}^\text{B}_\ell}{10pt}})$,
where $N_{\scaleto{\mathbf{s}^\text{A}_\ell}{10pt}} \cup N_{\scaleto{\mathbf{s}^\text{B}_\ell}{10pt}}$ denotes the set of the distinct packets passing  through at least one of nodes $\mathbf{s}^\text{A}_\ell$ and $\mathbf{s}^\text{B}_\ell$, and
set $N_{\scaleto{\mathbf{s}^\text{A}_1}{10pt}}\setminus (N_{\scaleto{\mathbf{s}^\text{A}_\ell}{10pt}} \cup N_{\scaleto{\mathbf{s}^\text{B}_\ell}{10pt}})$ represents 
the set of packets generated by source $\mathbf{s}^\text{A}_1$ but not received by node $\mathbf{s}^\text{A}_\ell$ or node $\mathbf{s}^\text{B}_\ell$.
Note that each node in a layer receives a mixture of some packets in traffic packets from both sources  $\mathbf{s}^\text{A}_1$ and $\mathbf{s}^\text{B}_1$.
Therefore, we can use the weighted Jaccard similarity $\mathcal{J_W}$ between traffic packets sets $N_{\scaleto{\mathbf{s}^\text{A}_\ell}{10pt}}$ and $N_{\scaleto{\mathbf{s}^\text{B}_\ell}{10pt}}$, i.e.,
$\mathcal{J_W}(N_{\scaleto{\mathbf{s}^\text{A}_l}{10pt}}, N_{\scaleto{\mathbf{s}^\text{B}_\ell}{10pt}})=\frac{|N_{\scaleto{\mathbf{s}^\text{A}_\ell}{10pt}}\cap N_{\scaleto{\mathbf{s}^\text{B}_\ell}{10pt}}|_{\text{w}}}{|N_{\scaleto{\mathbf{s}^\text{A}_\ell}{10pt}}\cup N_{\scaleto{\mathbf{s}^\text{B}_\ell}{10pt}}|_{\text{w}}}$,
to measure 
the proportion of the total size of identical packets passing through the two nodes.   
We show the results in Fig.~\ref{fig:re_sim4}. 
Given the Gumbel-Max sketches of two arbitrary sets $N_\mathbb{A}$ and $N_\mathbb{B}$,
Lemiesz~\cite{lemieszAlgebraDataSketches2021} proposed a series of methods to estimate 
the weighted cardinality of both union and intersection $|N_\mathbb{A}\cup N_\mathbb{B}|_\text{w}$ and $|N_\mathbb{A}\cap N_\mathbb{B}|_\text{w}$,
the weighted Jaccard similarity  $\mathcal{J_W}(N_\mathbb{A}, N_\mathbb{B})$,
the weighted cardinality of relative complement $|N_\mathbb{A}\setminus N_\mathbb{B}|_\text{w}$ from these sketches, and these methods can be extended to multiple sets.
In our experiments, we use the same methods to compute the total size of packets from sources A and B at each sensor node, the total size of lost packets at each sensor node, and the weighted Jaccard similarity between two nodes in each layer.
As we analyzed above, the $\vec{y}$ part of FastGM sketch is the same as Lemiesz's sketch, so they have the same performance in each experiment.  

To demonstrate the efficiency of our Stream-FastGM, in Fig.~\ref{fig:t2_simu_efc} we report the running time of generating sketches with different lengths $k$. When $k=2048$, our Stream-FastGM is $52$ times faster than Lemiesz's sketch, and the results show that our Stream-FastGM gets faster than Lemiesz's sketch when $k$ gets larger.
We also conduct experiments on simulated sensor networks with different depths of layers, as shown in Fig.~\ref{fig:t2_simu_efc_2}, our  Stream-FastGM is $47$ times faster than Lemiesz's sketch on average.

\section{Related Work} \label{sec:related}

\subsection{Jaccard Similarity Estimation}
Broder et al.~\cite{Broder2000} proposed the first sketch method \emph{MinHash} to
compute the Jaccard similarity of two sets (or binary vectors).
MinHash builds a sketch consisting of $k$ registers for each set.
Each register uses a hash function to keep track of the set's element with the minimal hash
value.
To further improve the performance of MinHash,
\cite{PingWWW2010,MitzenmacherWWW14,Wang2019mem} developed several
memory-efficient methods.
Li et al.~\cite{Linips2012} proposed \emph{One Permutation Hash} (OPH) to reduce the time complexity of processing each element from $O(k)$ to $O(1)$ but this method may exhibit large estimation errors because of the empty buckets.
To solve this problem, several densification
methods~\cite{ShrivastavaUAI2014,ShrivastavaICML2014,ShrivastavaICML2017,dahlgaard2017fast}
were developed to set the registers of empty buckets according to the values of
non-empty buckets' registers.

Besides binary vectors, a variety of methods have also been developed to estimate
generalized Jaccard similarity on weighted vectors.
For vectors consisting of only nonnegative integer weights, Haveliwala et
al.~\cite{haveliwala2000scalable} proposed to add a corresponding number of
replications of each element in order to apply the conventional MinHash.
To handle more general real weights, Haeupler et al.~\cite{haeupler2014consistent}
proposed to generate another additional replication with a probability that equals
the floating part of an element's weight.
These two algorithms are computationally intensive when computing hash values of
massive replications for elements with large weights.
To solve this problem, \cite{gollapudi2006exploiting, Manasse2010} proposed to
compute hash values only for a few necessary replications (i.e., "active
indices").
ICWS~\cite{IoffeICDM2010} and its variations such as 0-bit CWS~\cite{LiKDD2015},
CCWS~\cite{WuICDM2016}, PCWS~\cite{WuWWW2017}, I$^2$CWS~\cite{wu2018improved} were
proposed to improve the performance of CWS~\cite{Manasse2010}.
The CWS algorithm and its variants all have the time complexity of $O(n^+k)$,
where $n^+$ is the number of elements with positive weights.
Recently, Otmar~\cite{ertl2018bagminhash} proposed another efficient algorithm
\emph{BagMinHash} for handling high-dimensional vectors.
BagMinHash is faster than ICWS when the vector has a large number of positive
elements, e.g., $n^+>1,000$, which may not hold for many real-world datasets.
The above methods all estimate the weighted Jaccard similarity.
Ryan et al.~\cite{moulton2018maximally} proposed a Gumbel-Max Trick based
sketching method, $\mathcal{P}$-MinHash, to estimate another novel Jaccard
similarity metric, \emph{probability Jaccard similarity} $\mathcal{J_P}$.
They also demonstrated that the proposed probability Jaccard similarity $\mathcal{J_P}$ is scale-invariant
and more sensitive to changes in vectors.
However, the time complexity of $\mathcal{P}$-MinHash processing a weighted vector
is $O(n^+k)$, which is infeasible for high-dimensional vectors.

\subsection{Cardinality Estimation}
The regular problem of cardinality estimation aims to compute the number of distinct elements in the set of interest,
which is typically given as a sequence containing duplicated elements~\cite{lan2021survey}. 
To address this problem, 
a number of sketch methods such as LPC~\cite{whang1990linear}, LogLog~\cite{durand2003loglog}, HyperLogLog~\cite{flajolet2007hyperloglog},
RoughEstimator~\cite{kane2010optimal}, HLL-TailCut+~\cite{xiao2017better}, and HLL++~\cite{heule2013hyperloglog} build a sketch consisting of $m$ bits/counters for a set.
The sketch is small (e.g., $m=1,000$) and can be efficiently updated, which handles each element with few operations.
The generated sketch is finally used to estimate the set's cardinality.
In addition,~\cite{ting2014streamed,ertl2017new} exploit martingale estimation and maximum likelihood estimation to improve the estimation accuracy of the above methods.
For some applications, there may exist many sets of which sizes vary significantly.
To reduce the memory cost of building a sketch for each set,
a number of works~\cite{zhao2005joint,yoon2009fit,wang2011data,xiao2015hyper,wang2019utilizing,jia2020accurately,ting2019approximate} propose to implement $m$ independent hash functions to randomly map each sketch into a large shared bit/counter array,
where each sketch can be rebuilt by randomly sampling $m$ bits/counters from the shared array.

Recently, \cite{Cohen0Y15,lemieszAlgebraDataSketches2021} generalized the problem of cardinality estimation to a weighted version, where each element is associated with a fixed positive weight. 
The goal of weighted cardinality estimation is to estimate the total sum of weights for all distinct elements in the stream of interest.
The drawback of the sketch methods in~\cite{Cohen0Y15,lemieszAlgebraDataSketches2021} is their high computational costs.
\section{Conclusion} \label{sec:conclusions}
In this paper, we develop an efficient algorithm \emph{FastGM}  to compute a non-negative vector's $k$-length Gumbel-Max sketch.
We propose a novel model, \emph{Queuing model with $k$-servers and $n$-queues}, to model the procedure of computing the Gumbel-Max sketch in a brief and practical way.
Based on the proposed model, we optimize the procedure of generating $k$ random variables $-\frac{\ln a_{i,j}}{v_i}$ of an element $v_i$ in a vector.
We theoretically prove that our \emph{FastGM} reduces the time complexity of generating a $k$-length Gumbel-Max sketch from $O(n^+ k)$ to $O(k \ln k + n^+)$, where $n^+$ is the number of the vector's positive elements.
We conduct two tasks probability Jaccard similarity estimation and weighted cardinality estimation to demonstrate the efficiency and effectiveness of \emph{FastGM}.
Experimental results show that our \emph{FastGM} is around $10$ times faster than state-of-the-art methods, without losing any estimation accuracy.

% use section* for acknowledgment
%\ifCLASSOPTIONcompsoc
  % The Computer Society usually uses the plural form
\section*{Acknowledgments}
%\else
  % regular IEEE prefers the singular form
%  \section*{Acknowledgment}
%\fi

The authors would like to thank the anonymous reviewers for their comments and suggestions.
% This work was supported in part by the National Key R\&D Program of China
% (2021YFB1715600), National Natural Science Foundation of
% China (61902305, 61922067), Shenzhen Basic Research Grant
% (JCYJ20170816100819428), MoE-CMCC "Artificial Intelligence" Project (MCM20190701).
This work was supported in part by 
National Natural Science Foundation
of China (U22B2019, 62272372, 61902305), 
MoE-CMCC "Artificial Intelligence" Project (MCM20190701).

% Can use something like this to put references on a page
% by themselves when using endfloat and the captionsoff option.
%\ifCLASSOPTIONcaptionsoff
%%\fi

\bibliographystyle{IEEEtran}
\bibliography{COPH,randpe,ctstream,simcar,albitmap,dynamic,Imfastgm,references}

\begin{IEEEbiography}  
[{\includegraphics[width=1in,height=1.25in,clip,keepaspectratio]{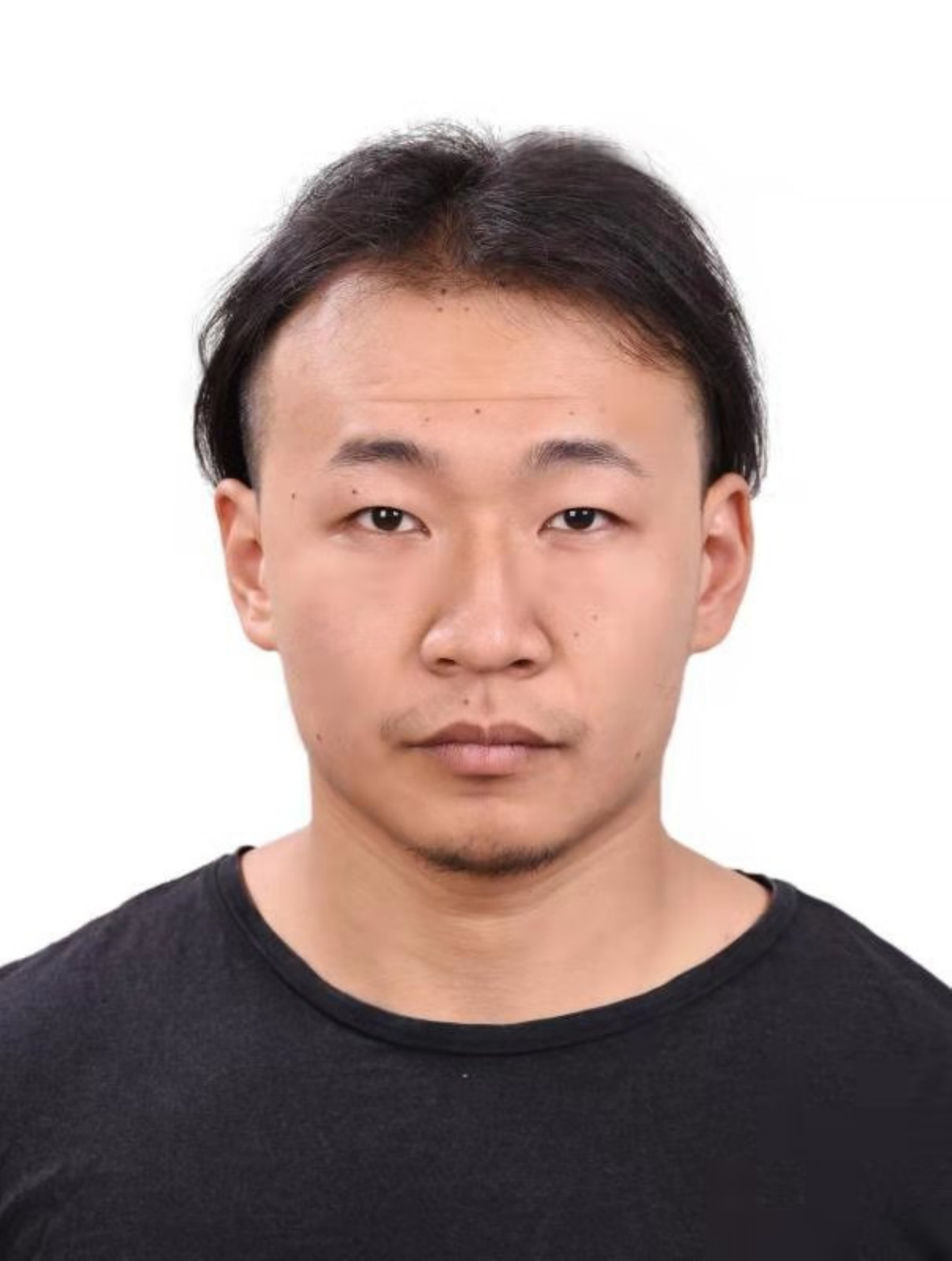}}] {Yuanming Zhang} received a B.S. degree in automation from Chongqing University, Chongqing, China, in 2017. He is currently working toward a graduate degree at the MOE Key Laboratory for Intelligent Networks and Network Security, Xi'an Jiaotong University, Xi'an, China. His research interests include anomaly detection, encrypted traffic analysis, and Internet traffic measurement and modeling. 
\end{IEEEbiography}

\begin{IEEEbiography}
[{\includegraphics[width=1in,height=1.25in,clip,keepaspectratio]{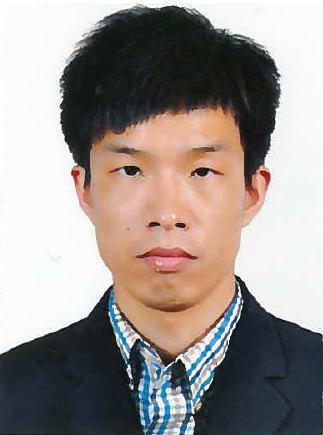}}]
{Pinghui Wang} (Senior Member, IEEE) is currently a Professor with the MOE Key Laboratory for Intelligent Networks and Network Security, Xi'an Jiaotong University, Xi'an, China, and also with the Shenzhen Research Institute, Xi'an Jiaotong University, Shenzhen, China. His research interests include internet traffic measurement and modeling, traffic classification, abnormal detection, and online social network measurement
\end{IEEEbiography}

\begin{IEEEbiography}  
[{\includegraphics[width=1in,height=1.25in,clip,keepaspectratio]{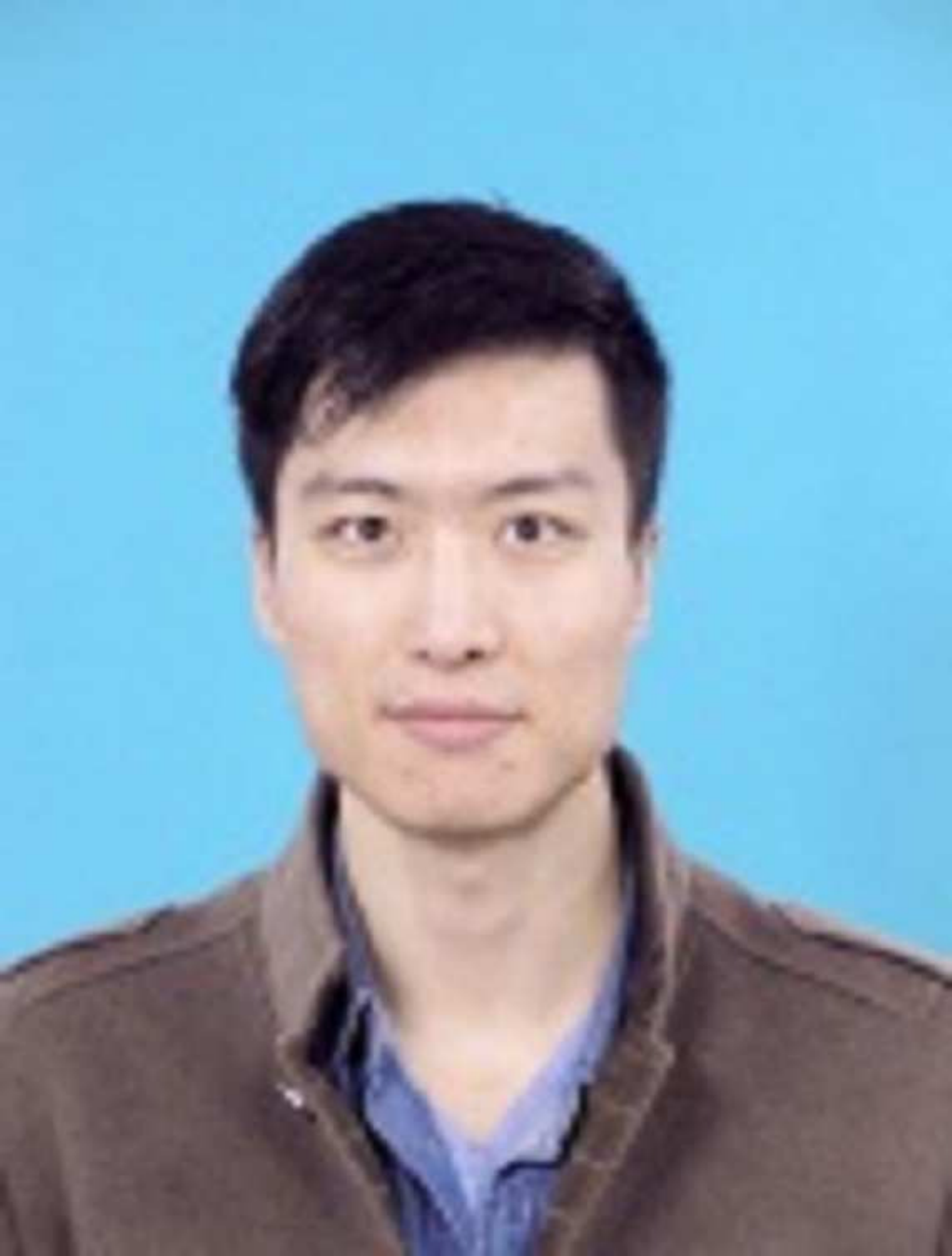}}] {Yiyan Qi} received a B.S. in automation engineering and a Ph.D. degree in automatic control from Xi'an Jiaotong University, Xi'an, China, in 2014 and 2021 respectively. He is currently a Researcher at the International Digital Economy Academy (IDEA). Prior to joining IDEA, he was working at Tencent. His current research interests include abnormal detection, graph mining and embedding, and recommender systems.
\end{IEEEbiography}

\begin{IEEEbiography}  
[{\includegraphics[width=1in,height=1.25in,clip,keepaspectratio]{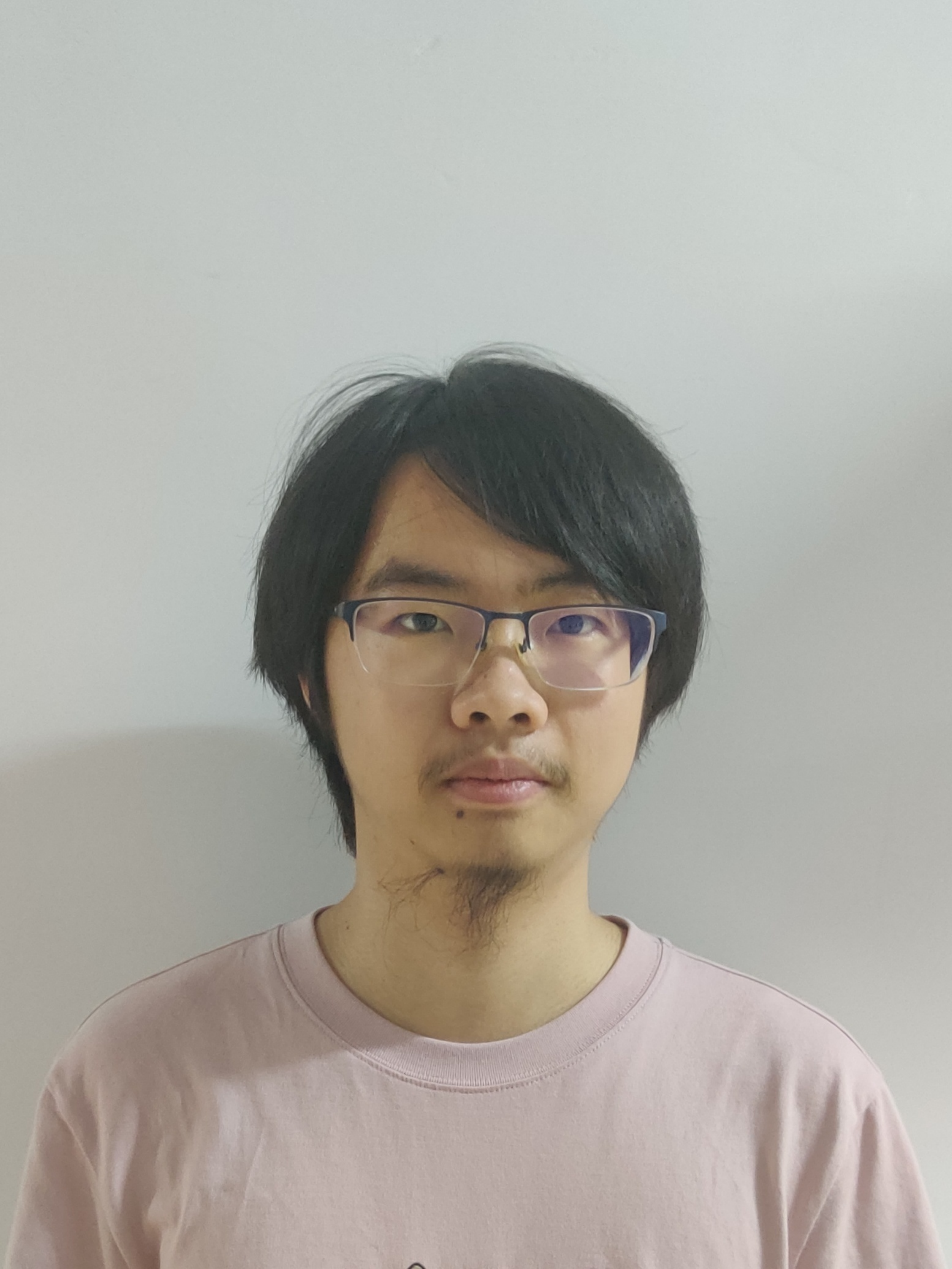}}] {Kuankuan Cheng} is currently working toward an undergraduate degree at Xi'an Jiaotong University, Xi'an, China. His research interests include streaming data processing and encrypted traffic analysis.
\end{IEEEbiography}

\begin{IEEEbiography}
[{\includegraphics[width=1in,height=1.25in,clip,keepaspectratio]{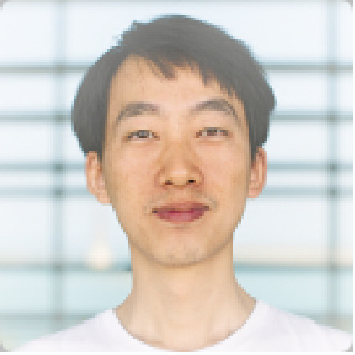}}]
{Junzhou Zhao} received B.S. (2008) and
Ph.D. (2015) degrees in control science and engineering from Xi'an Jiaotong University. He is
currently an associate professor at the School
of Cyber Science and Engineering, Xi'an Jiaotong University. His research interests include
graph data mining and streaming data processing.
\end{IEEEbiography}

\begin{IEEEbiography}
[{\includegraphics[width=1in,height=1.25in,clip,keepaspectratio]{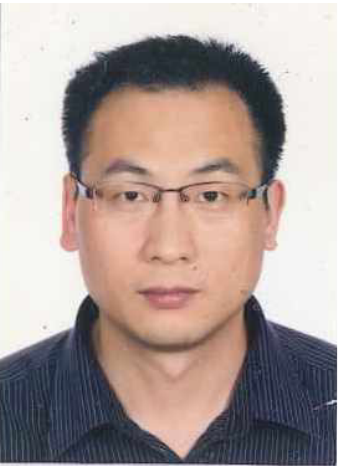}}]
{Guangjian Tian} received a Ph.D. degree in computer science and technology from Northwestern Polytechnical University, Xi'an, China, in 2006. He is currently a principal researcher in Huawei Noah's Ark Lab. Before that, he was a postdoctoral research fellow with the Department of Electronic and Information Engineering, The Hong Kong Polytechnic University, Hong Kong. His research interests include temporal data analysis, deep learning, and data mining with a specific focus on different industry applications.
\end{IEEEbiography}

\begin{IEEEbiography}
[{\includegraphics[width=1in,height=1.25in,clip,keepaspectratio]{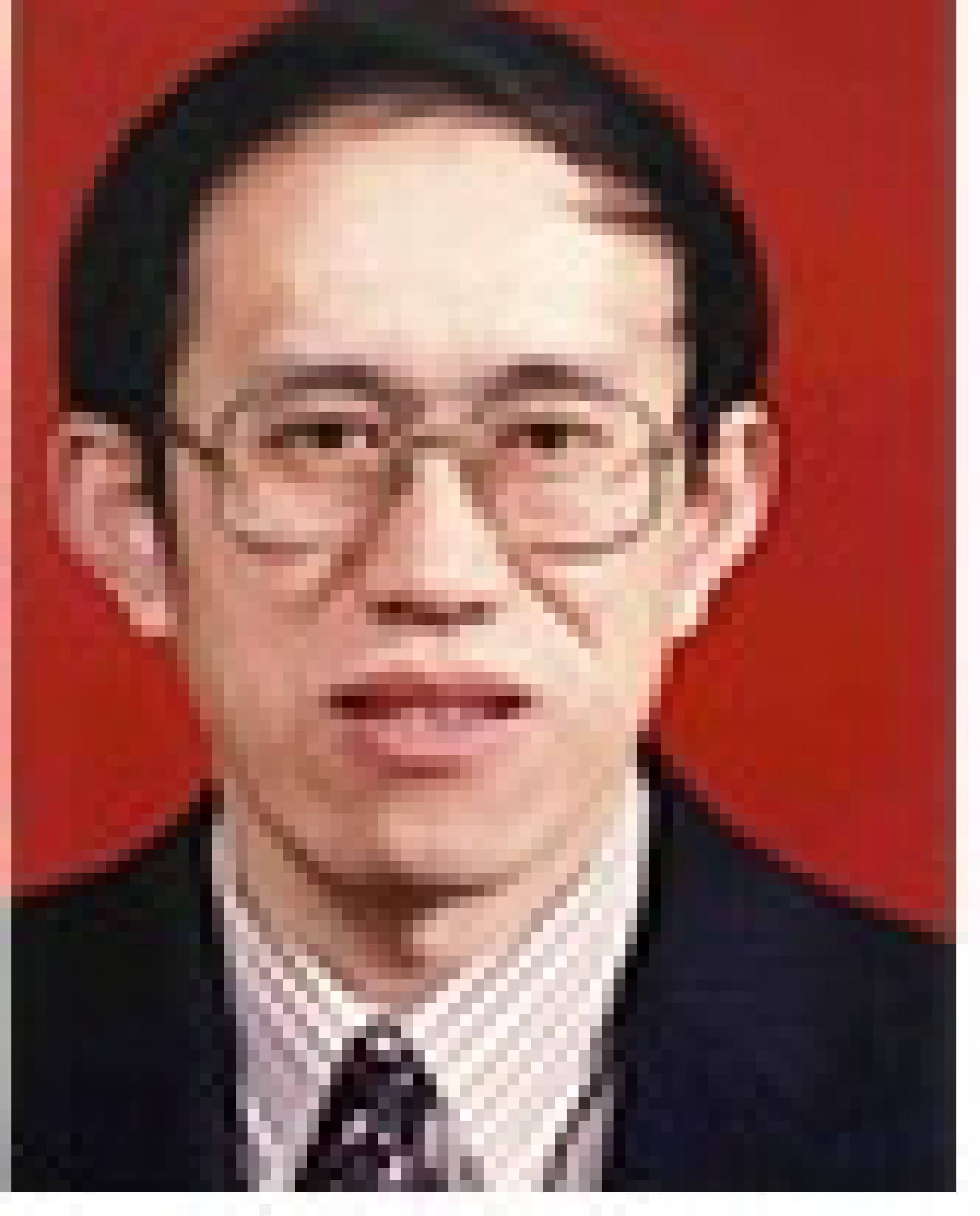}}]
{Xiaohong Guan} (Fellow, IEEE) received a Ph.D. degree in electrical engineering from the University of Connecticut, Storrs, in 1993. Since 1995, he has been with the Department of Automation, Tsinghua National Laboratory for Information Science and Technology, and the Center for Intelligent and Networked Systems, Tsinghua University.
He is currently with the MOE Key Laboratory for Intelligent Networks and Network Security, Faculty of Electronic and Information Engineering, Xi'an Jiaotong University, Xi'an, China, where he is also the Dean of the Faculty of Electronic and Information Engineering.
He is an Academician of the Chinese Academy of Sciences.
\end{IEEEbiography}

\end{document}